\documentclass[journal,10pt,onecolumn,draftclsnofoot,]{IEEEtran}

\ifCLASSINFOpdf
\usepackage[pdftex]{graphicx}
\graphicspath{{/figs/}}

\DeclareGraphicsExtensions{.pdf,.eps}
\else

\usepackage[dvips]{graphicx}

\graphicspath{{/figs/}}

\DeclareGraphicsExtensions{.eps}
\fi
\usepackage{xspace}
\usepackage{caption}
\usepackage{subcaption}
\usepackage{xcolor}
\usepackage[cmex10]{amsmath}
\usepackage{mathtools,amssymb,lipsum}
\usepackage{lettrine}
\usepackage{cite}
\usepackage{bm}
\usepackage{mathrsfs}
\usepackage{algorithm,tabularx} 
\usepackage{algpseudocode}
\makeatletter
\newcommand{\multilines}[1]{%
	\begin{tabularx}{\dimexpr\linewidth-\ALG@thistlm}[t]{@{}X@{}}
		#1
	\end{tabularx}
}
\makeatother
\usepackage{array}
\usepackage{url}  
\usepackage{amsthm}
\usepackage{multirow}
\usepackage{color}
\usepackage{epstopdf}
\usepackage{endnotes}
\usepackage{framed}
\usepackage{cool} 
\usepackage{enumerate} 
\usepackage{url}
\usepackage[subnum]{cases}
\usepackage[nocomma]{optidef}
\usepackage{etoolbox}

\newtheorem{definition}{Definition}

\makeatletter
\newcommand{\multiline}[1]{%
  \begin{tabularx}{\dimexpr\linewidth-\ALG@thistlm}[t]{@{}X@{}}
    #1
  \end{tabularx}
}

\newtheorem{remark}{Remark}

\makeatletter
\patchcmd{\@maketitle}
{\addvspace{0.5\baselineskip}\egroup}
{\addvspace{-2\baselineskip}\egroup}
{}
{}
\makeatother

\begin{document}
\title{Edge-assisted Democratized Learning Towards Federated Analytics}
\author{Shashi~Raj~Pandey,~\IEEEmembership{Student Member,~IEEE}\thanks{S.R. Pandey, M.N.H Nguyen, T.N. Dang, K. Thar, and C.S Hong are with the Department of Computer Science and Engineering, Kyung Hee University, Yongin 17104, South Korea.},~Minh~N.~H.~Nguyen,~\IEEEmembership{Member,~IEEE}, ~Tri~Nguyen~Dang,\\~Nguyen~H.~Tran,\thanks{N. H. Tran is with School of Computer Science, The University of Sydney,
Sydney, NSW 2006, Australia.}~\IEEEmembership{Senior Member,~IEEE},~Kyi~Thar,~Zhu~Han\thanks{Z. Han is with the Electrical and Computer Engineering Department, University of Houston, Houston, TX 77004, USA, and also with the Department of Computer Science and Engineering, Kyung Hee University, Seoul, South Korea.},~\IEEEmembership{Fellow,~IEEE,}~Choong~Seon~Hong,~\IEEEmembership{Senior Member,~IEEE}}
\maketitle

\begin{abstract}
A recent take towards \textit{Federated Analytics} (FA), which allows analytical insights of distributed datasets, reuses the Federated Learning (FL) infrastructure to evaluate the summary of model performances across the training devices. However, the current realization of FL adopts single server-multiple client architecture with limited scope for FA, which often results in learning models with poor generalization, i.e., an ability to handle new/unseen data, for real-world applications. Moreover, a hierarchical FL structure with distributed computing platforms demonstrates incoherent model performances at different aggregation levels. Therefore, we need to design a robust learning mechanism than the FL that (i) unleashes a viable infrastructure for FA and (ii) trains learning models with better generalization capability. In this work, we adopt the novel democratized learning (Dem-AI) principles and designs to meet these objectives. Firstly, we show the hierarchical learning structure of the proposed edge-assisted democratized learning mechanism, namely \textit{Edge-DemLearn}, as a practical framework to empower generalization capability in support of FA. Secondly, we validate Edge-DemLearn as a flexible model training mechanism to build a distributed control and aggregation methodology in regions by leveraging the distributed computing infrastructure. The distributed edge computing servers construct regional models, minimize the communication loads, and ensure distributed data analytic application's scalability. To that end, we adhere to a near-optimal two-sided many-to-one matching approach to handle the combinatorial constraints in Edge-DemLearn and solve it for fast knowledge acquisition with optimization of resource allocation and associations between multiple servers and devices. Extensive simulation results on real datasets demonstrate the effectiveness of the proposed methods.
\end{abstract}
	

\IEEEpeerreviewmaketitle
	
\section{Introduction}\label{sec: introduction}

\subsection{Background and Motivation}
Recently, the Google AI team published an article on leveraging computing mechanism of the distributed learning model training infrastructure to facilitate data analytics, namely \textit{Federated Analytics} (FA) \cite{federatedanalytics}. FA allows data scientists to derive analytical insights of distributed datasets without the need of moving data to a central computing entity. This concept gathered keen attention for a new approach to data science, and interestingly, at the time when the centralized repositories are termed ``vulnerable" towards privacy for data collection,  and the era of distributed computing and storage is prominent. Besides, it means that apart from considering the distributed model training processes for improving model accuracy, we can exploit such collaboration architecture to evaluate the quality of the trained model at the user-level perspectives, i.e., the model performance at the user's end. Hence, \textit{without the learning part}, we can reuse the computing scheme of the learning architecture to perform statistical analysis on local data that may lead to building better products. To elaborate this idea further, consider an example of a prediction model where the developer would be interested in finding popular contents to store in a shared regional database without breaking into user's historical content usage data. An intuitive answer to this question would be to find the frequently requested content at first, which is best done with FA. This is similar to the \textit{Now Playing} feature on Google's Pixel phones for managing regional song database \cite{federatedanalytics} to show users songs playing around them. In addition, leveraging distributed multi-access edge computing (MEC) servers further allows fast knowledge acquisition to serve user's requests, limit privacy leakages, and improve the \textit{on-the-fly} learning process. Furthermore, this also brings FA closer to where the data is collected, i.e., at the one-hop proximity of user devices. In such scenarios, the developer's objective would be to improve the model's accuracy and, concurrently, enhance the generalization performance of the model at the user level for maintaining regional databases using FA.

In this regard, the ubiquitous presence of edge computing devices such as mobile phones and sensors has unleashed the potential of exploiting distributed data for several learning applications, namely \textit{edge learning} \cite{zhu2020toward}. This has opened up further opportunities to renovate the edge learning architectures for FA. Besides, the improvement in computational capabilities of the next-generation mobile devices using systems-on-chip (SoC) \cite{atabaki2018integrating} has helped leverage distributed on-device model training mechanism to realize different learning schemes. Nevertheless, there are concerns related to privacy issues to fully tap the benefits of distributed data and perform statistical analysis upon them for various services, e.g., regional  database  management \cite{federatedanalytics}. To that end, FA can work together with federated learning (FL), which is a promising paradigm to address privacy concerns during model training \cite{mcmahan2017communication, li2020federated, pandey2020crowdsourcing, FL_advances, khan2020federated} and supports with the computing infrastructure for FA \cite{federatedanalytics}. However, the current rigid topology of the FL architecture in wireless networks \cite{pandey2020crowdsourcing, chen2020convergence, tran2019federated, le2020incentive} hinders the full potential of exploiting FA for next-generation of personalized applications. In particular, the typical realization of FL over wireless adopts a straightforward single server-multiple client architecture with limited scope for FA. Moreover, a hierarchical FL structure with distributed computing platforms results in incoherent model performances at different aggregation levels.

Additionally, besides FL infrastructure being considered an appropriate framework for FA, another practical issue for FL over the wireless environment is the channel variations. Hence, the communication cost between the devices and the MEC server strongly affects the learning performance \cite{FL_advances, amiri2020federated, tran2019federated}, leading poor model quality. Furthermore, the number of UEs involved in the distributed model training is another critical factor that affects the quality of  the trained global model and learning performance \cite{nishio2019client, pandey2020crowdsourcing}. Therefore, it is imperative to efficiently allocate limited wireless resources while taking care of devices' local computing strategies to enhance communication efficiency further, and meet the requirements for FA as mentioned above. Also, several inherent issues, such as statistical and system-level heterogeneity, scalability, highly personalized and unbalanced data triggers the most critical challenge of vanilla FL implementation over wireless networks: improving both the learning performance at the user's end, namely \emph{personalized learning} performance, and an ability to handle new/unseen data, namely the \emph{generalization capability}, for the real-world applications such as in healthcare, robotics, and autonomous vehicles. However, the current implementation of FL denounces user-level heterogeneity. It restricts users from forming groups and collaborating as per the characteristics of their learning tasks, which affects personalization. Further, this scenario also limits efficient knowledge transfer, which is in terms of model parameters aggregation at a different aggregation level, and \textit{knowledge acquisition time}. We term such scenario as the inability of \textit{self-organization}.

In this regard, several works tackle above challenges under two broader domains to meet \textit{communication-efficient} and \textit{energy-efficient} FL implementations, while leaving aside the expected robust framework for FA: i) resource management in radio access networks and ii) client selection. For example, the authors in \cite{yang2020federated} implemented an over-the-air computation approach to facilitate fast aggregation via joint device selection and beamforming design. Similarly, the authors in \cite{tran2019federated} evaluated various trade-offs in FL schemes over wireless networks involving the uncertainty of wireless channels and heterogeneous participating devices. In particular, \cite{tran2019federated} explored and optimized the inter-dependencies between the computation-communication latency and learning time-energy consumption constraints for improving learning performance. However, the underlying network architecture is \textit{traditional} (a single server and multiple devices collaboratively training a single model, and data homogeneity is considered), similar to several existing works \cite{pandey2020crowdsourcing, chen2019joint, le2020incentive, yang2020federated}, and haven't analyzed hierarchical knowledge transfer at different network points, such as edge nodes. Thus, this limits the involvement of edge infrastructure to support hierarchical knowledge transfer and minimizes the model training time. Also, the single trained model, naively imposed on devices in each communication round, is poor in generalization, resulting in larger generalization errors \cite{federatedSurvey2020, FL_advances}. Therefore, it is necessary to have a distributed learning mechanism that considers the personal attributes of participating devices and results in tailored individual models while collaborating to train a global model with better generalization abilities.

Following the above discussions, we come to two primary, yet overlooked research questions on the distributed learning infrastructure in support of FA:
\begin{itemize}
    \item \textit{How to unleash a viable infrastructure for FA other than FL leveraging distributed MEC infrastructures at different regional levels when designing distributed model training architecture?}
    \item \textit{How to realize a more practical, robust and flexible learning structure towards improving generalization that accommodates heterogeneous learning UEs, results fast knowledge acquisition time, and ensures better FA?}
\end{itemize}

We resort to a self-organizing  hierarchical  learning  structure that is resilient  to  generalization  errors  while  improving personalization, and the best-suited for FA at different levels. In  particular, we adopt novel Dem-AI  principles \cite{nguyen2021distributed}  and  design a  distributed  control and  aggregation  methodology  in  regions by  leveraging  distributed  MEC platforms in a heterogeneous network (HetNet) architecture with multiple small cell eNodeBs (SBSs) and a macro cell eNodeB (MBS). Moreover, we exploit a low-complexity two-sided many-to-one matching solution \cite{gale1962college, roth1992two} to realize interaction amongst UEs, and resolve association between the SBSs and the UEs to meet the objectives of FA while implementing edge-assisted Dem-AI.

To that end, in the following subsection, we first briefly introduce the learning mechanism in Dem-AI systems, and the resulting hierarchical structure for FA, followed by contributions, and finally the organization of this paper. 

\subsection{Preliminaries: Democratized Learning Systems}\label{i_b}
   The Dem-AI system \cite{nguyen2021distributed} aims to ensure \emph{democracy in a distributed learning system} as follows:
    \begin{itemize}
    \item learning UEs are self-organized into appropriate hierarchical groups according to their learning characteristics. This process mediates \emph{contributions} from all members in the collaborative learning to build corresponding hierarchical generalized knowledge. 
    \item The shared hierarchical generalized learning knowledge supports UEs to speed up their personalized learning process and contribute back enhance the generalization capability of groups' knowledge.
    \end{itemize}

The Dem-AI philosophy challenges the consistency in doing regional training and aggregation to perform knowledge transfer. The general Dem-AI concepts and guidelines in \cite{nguyen2021distributed} inspire us to develop our initial implementation of Dem-AI in \cite{nguyen2020self}. Moreover, the self-organized hierarchical learning structure exposes a new approach of doing FA, particularly, at different levels and regions. Therefore, beyond model evaluation, FA enables to better handle regional databases leveraging the the proposed learning infrastructure. 

\subsection{Contributions}\label{i_c}
In this work, we design and analyze a novel edge-assisted distributed learning paradigm that offers FA services at different levels, while delivering high-quality learning models. In particular, we propose a more robust and flexible model training mechanism that results in a hierarchical architecture, which is the best-suited for meeting both the FA and learning objectives. Specifically, the main contributions of the paper are summarized as follows:
\begin{itemize}
    \item We introduce a novel edge-assisted hierarchical learning structure, namely \textit{Edge-DemLearn}, which acts as a practical framework to empower generalization capability in support of FA. We show Edge-DemLearn characterizes the quality of global model as the prediction performance at UEs, and at different regional levels, for FA. 
    \item We formulate an optimization problem to handle joint decision on users association and network resource allocations, i.e., communication and computing resources, for exchanging learning models in Edge-DemLearn. The problem minimizes the overall delay of the aggregation operation process to speed up the regional learning process.
    \item We apply a sub-optimal two-sided many-to-one matching algorithm to handle combinatorial constraints in Edge-DemLearn, and optimize the association between the number of UEs participating in the model training process and the SBSs. We leverage multi-connectivity scenario in the HetNet topology and obtain a stable matching outcome to continue the model training process. 
    \item At the same time, we develop hierarchical clustering algorithm to arrange UEs in logical groups and then perform model aggregation at different levels, different than hierarchical federated learning (HFL).
    \item Extensive simulations performed on well-known datasets, MNIST, FEMNIST, and Fashion-MNIST, demonstrate the learning performance (i.e., 95\% generalization) of the proposed approach in a 2-layer HetNet architecture. Simulation results depict significant gain in system performance of 26.2\% as compared to the baselines; and hence, ensuring low-latency model aggregation or fast knowledge acquisition.
\end{itemize}

\subsection{Organization}\label{i_d}
The rest of the paper is organized as follows. Section \ref{sec: related_works} reviews the related works, discussing several approaches in handling issues related with distributed learning over wireless networks and its implementation challenges. Section \ref{sec:system_model} introduces the system model of edge-assisted democratized learning model and a general problem formulation for fast-knowledge acquisition. Section \ref{sec:framework} presents Dem-AI systems at the network edge in support of FA at different levels, problem formulation for joint association and resource allocation design in heterogeneous networks for minimizing overall learning time and better generalization, and develops and analyse a two-sided many-to-one matching solution for the proposed mechanism. Section \ref{sec: Solution Approach} provides the performance evaluation of the proposed approaches and compares with other state-of-the-art approaches using real-world datasets. Finally, Section \ref{sec:performance_evaluation} concludes this work.

\section{Related Works}\label{sec: related_works}
We first review the related works on distributed machine learning techniques and discuss several performance issues related with it\footnote{Here, we focus more on the performance of FL, it limitations and challenges due to the inherent architecture, to evaluate the adaptability of FA.}. In particular, we discuss a typical FL setting, several related algorithms, and its implementation challenges over wireless networks such as intermittent network connectivity, statistical and system-level heterogeneity, resource allocation and participation, convergence guarantees, and personalization-generalization errors. Second, we review available solution approaches to realize FL over wireless networks. Basically, we discuss related works on handling communication bottleneck and delays, resource allocation strategies (i.e., computing and communication resources), participants selection, system-level heterogeneity, and convergence time minimization in doing so for several network architectures.   

In the FL scheme, the local dataset collected at each UE is unbalanced and highly personalized, which reflects the personality of each UE (or users). Adopting a simple model averaging scheme in FedAvg \cite{mcmahan2017communication} or FedProx \cite{li2020federated} using local parameters of learning UEs who have exceedingly different characteristics may result in poor personalized performance. In this regard, on the one hand, recent works in \cite{fallah2020personalized, deng2020adaptiveFL} made the first attempt to study the personalization in FL, however, the correlation between generalization and personalization was not adequately analyzed. On the other hand, algorithms like FedAvg show poor performance in a wireless environment, and in particular, due to the challenges posed by statistical heterogeneity, i.e., non-i.i.d. (independent and identical distribution) and unbalanced dataset, and system-level heterogeneity on the accuracy of trained model and the convergence rate \cite{FL_advances}. This is in line to the findings when adopting several other distributed learning algorithms such as Parallel-Stochastic Gradient Descent (P-SGD) \cite{zinkevich2010parallelized} and Decentralized-Parallel SGD (D-PSGD) \cite{lian2017can} as well. In a general distributed setting like FL, where the UEs frequently interact with a central aggregator, such algorithms are vulnerable in the centralized wireless setting due to untimely or delayed parameter exchanges, requiring larger communication rounds between the UEs and the aggregator; and hence, resulting poor communication efficiency. In order to improve the communication efficiency, techniques of quantization (e.g., signSGD \cite{bernstein2018signsgd}, Q-SGD\cite{alistarh2017qsgd}), compression \cite{lin2017deep}, sampling \cite{konevcny2016federated}, and sparsification \cite{sattler2019robust} have been studied. In this regards, the authors in \cite{konevcny2016federated} discussed about a family of new randomized methods combining SGD, with primal and
dual variants such as Stochastic Variance Reduced Gradient (SVRG), Federated Stochastic Variance Reduced Gradient (FSVRG) and Stochastic Dual Coordinate Ascent (SDCA). More recently, the authors in \cite{wang2018edge} analyzed the convergence rate of the distributed gradient descent considering resource constrained edge computing system. Considerably, all of these works aim to enable communication-efficient FL, and further study the convergence properties.
Apart from techniques such as model compression, quantization, and coding to reduce the communication overheads required in distributed learning, optimal scheduling and radio resource distribution amongst UEs in a wireless setting is imperative for its practical implementation. In this regard, the authors in \cite{shi2019device} proposed a joint device scheduling and resource allocation to minimize overall model training time. Similarly, in \cite{chen2019joint} and \cite{chen2020convergence}, the authors focused on minimizing learning time while improving communication efficiency to realize FL over wireless networks. Here, the authors considered a general network architecture where multiple devices are connected to a single base station (BS) for model parameter aggregation. Similarly, several works such as \cite{tran2019federated}, \cite{zeng2019energy} have also considered the energy consumption while optimizing learning performance in the distributed machine learning paradigm of FL. In particular, optimal local computations and communication metrics are derived to solve an optimization problem defined as the weighted sum of learning time and the energy consumption. However, the outstanding issue related with generalization and specialization is overlooked. To that end, a few works \cite{abad2020hierarchical}, \cite{luo2020hfel} have discussed a hierarchical structure in realizing FL model training. The authors in \cite{abad2020hierarchical} presented a hierarchical federated learning (HFL) scheme to improve communication latency without compromising the achievable accuracy of the trained model. Similarly, the work in \cite{luo2020hfel}, which is closely related to ours, the authors considered an edge-cloud hierarchical architecture to perform model aggregation and proposed an energy-efficient resource scheduling algorithm to obtain model training performance gain compared to the vanilla FL setting. Unlike this work, we have considered multi-level operation leveraging flexible model aggregation framework for regional model construction in a HetNet environment for FA. In particular, under our proposed mechanism, the devices group amongst themselves to improve their specialization abilities, minimize generalization errors via knowledge transfer at different hierarchical layers and obtain higher performance gain for the trained global model. Furthermore, we focus more on improving knowledge acquisition period leveraging MEC servers at the network edges following Dem-AI principles, rather than just considering users scheduling in the proposed network architecture.   

\section{System Model}\label{sec:system_model}
We consider a multi-connectivity scenario \cite{simsek2019multiconnectivity} for multi-path wireless network links and adopt a 2-layer heterogeneous network (HetNet) topology with a single macrocell eNodeB (MBS) at layer 2 and a set of overlaid small cell eNodeBs (SBSs) $\mathcal{S}$ of $|\mathcal{S}| = S$ at layer 1. The SBSs are placed randomly within the coverage of MBS. We define a set of user equipments (UEs) $\mathcal{N}$ of $|\mathcal{N}| = N$ randomly placed within the cellular network. Furthermore, we also define a set of logical learning learning groups, $\mathcal{G} = \{1,2, \ldots,G\}$, in the considered region with a set of UEs $\mathcal{N}_{g\in \mathcal{G}}= \{N_1,N_2,\ldots,N_g\}$ associated with each group $g \in \mathcal{G}$ for training the regional learning model. The logical learning groups are formed based on the similarities in learning characteristics of the UEs, such as learning parameters (e.g., model weights and gradients) and aims at improving their group knowledge via the distributed model training approach. Then, considering $\mathcal{N}_g^s$ as the number of UEs belonging to the logical learning group $g$ and $\mathcal{N}^s$ as the number of UEs associated with SBS $s$, we have $\mathcal{N}^s =\cup_{g \in \mathcal{G}}\mathcal{N}_g^s$.
\begin{figure*}[t!]
	\centering
	\includegraphics[width=0.65\linewidth]{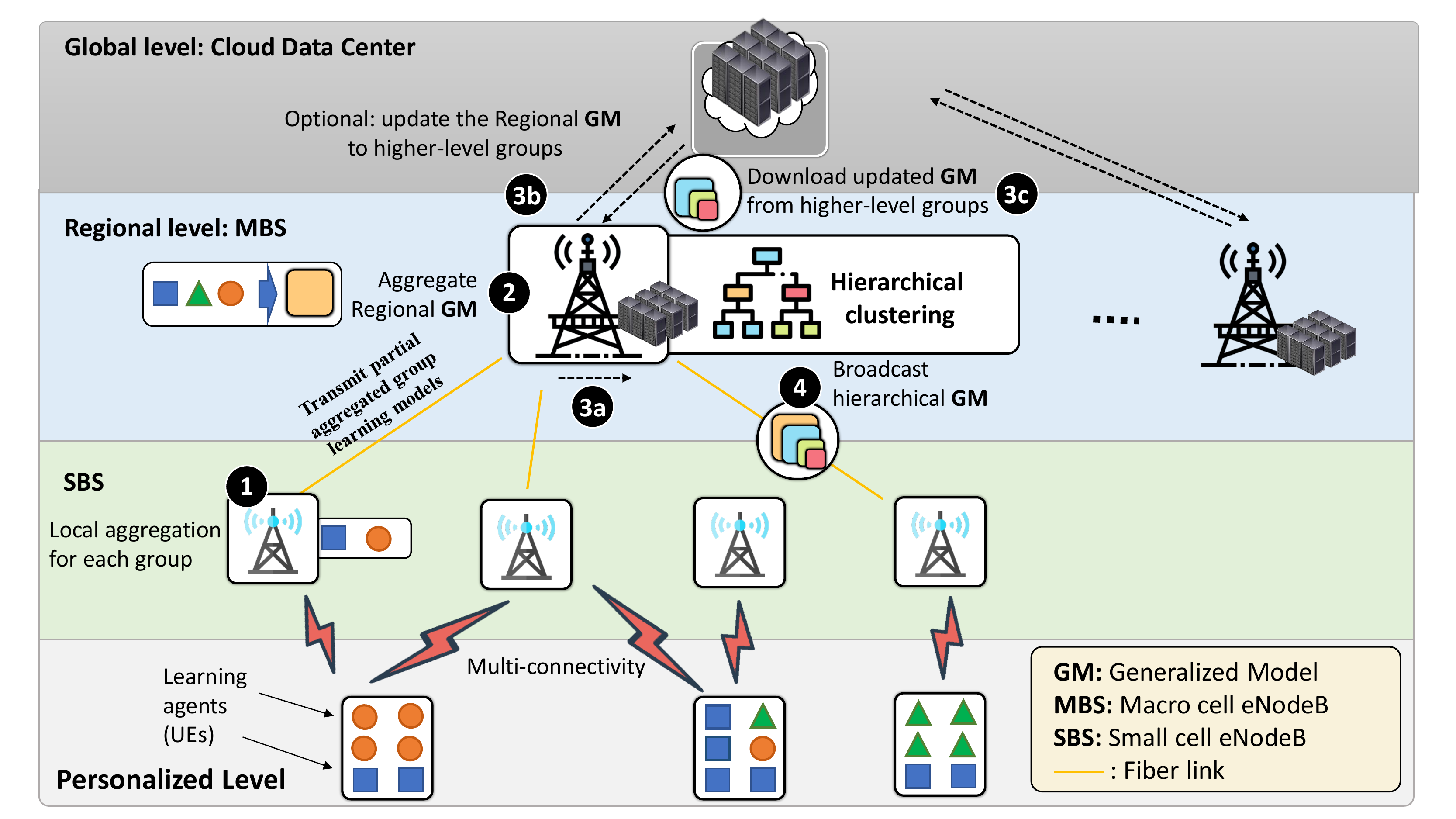}
	\caption{Proposed edge-assisted democratized learning model: a generalized view.}
	\label{F:Edge_DemAI}
\end{figure*}

Furthermore, according to the 3GPP standard \cite{3gpp_release15}, we consider that MBS and SBSs operate in adjacent sub 6-GHz  frequencies, where we assume the total bandwidth available at the SBSs, operating at the same carrier frequency, is $B$. This bandwidth is later shared amongst the associated UEs of the logical learning groups using an orthogonal frequency division multiplexing (OFDMA) technique. In particular, for a given number of $\Delta$ sub-bands, the available bandwidth $B$ is equally divided into several resource blocks (RBs). Therefore, each RB, which is defined as a sub-band in SBS with the bandwidth of $\hat{M}_{RB} = \frac{B}{\Delta}$. Thereby, the total number of resource blocks (RBs) is $M_{RB} = S \times\Delta$. Furthermore, in a multi-connectivity scenario, UEs can be associated with one or more SBSs based on the reference signal received power (RSRP) \cite{simsek2019multiconnectivity}. Furthermore, UEs can also change their associations with the SBSs following a certain rule. Hence, we draw a conclusion that multi-connectivity allows UEs to execute propose-and-reject algorithms following a matching game so as to improve their utility \cite{gale1962college}, \cite{roth1992two}. In what follows, the MBS manages the UEs control plane and the data transmission is accomplished by the SBSs. Table~\ref{tab:table1} provides the summary of key notations used in this work. \par

\begin{table}[t!]
	\centering
	\caption{Summary of key notations.}
	\label{tab:table1}
	\begin{tabular}{ll}
		\hline
		
		Notation & Definition\\
		\hline
            $\mathcal{S}, \mathcal{N}, \mathcal{G}$ & Set of SBSs, user equipments (UEs),\\ 
            & and logical learning groups, respectively \\ 
            
			$\mathcal{N}^s_g$ & Set of UEs associated with SBS $s$ and\\
			& belonging to logical learning group $g$ \\
			
			$B, \Delta, \hat{M}_{RB}$ & Available bandwidth, the  number of sub-bands,\\
			& and the bandwidth of each RB, respectively \\ 
			$\xi_{n, s}$ & Number of RBs assigned to the UE $n$ from SBS $s$\\
			
			$ r_{n, s}$ & Achievable throughput of user $n$ from SBS $s$\\
			
			$ r_{n, \tilde{S}_n}$ & Overall throughput when the UE $n$ is assigned to \\
			& a set of SBSs as $\tilde{S}_n \subseteq \mathcal{S}$\\
			
			$\beta_n^s$ & Fraction of bandwidth allocation for device $n$\\ & from SBS $s$ \\
			
			$w$ & Model parameter during one global iteration \\			
			$D_n^g, D^g$ & Size of local dataset of UE $n$ belonging to group $g$, \\
			& and total training data size of group $g$, respectively\\
			
			$J_{n}(w|\mathcal{D}^g_n) $ &  General empirical loss with respect to $w$ on the \\
    		& local data set $\mathcal{D}_n$ \\ 
    		
			$J_g(w)$ &  Learning loss function of group $g$ \\
			
			$I^{\textrm{global}}(\epsilon, \theta)$ & Number of global iterations \\
			
			$ I^{\textrm{local}}(\theta)$ & Number of local iterations\\
			$t_n^{\textrm{comp}}$ & Total computation time per global iteration\\ & at each UE $n$\\
			$c_n, f_n $ & Total number of CPU-cycles, and allocated \\& CPU-frequency of UE $n$, respectively.\\
			
			$t_{n,s}^{\textrm{com}}$ &  Latency during parameter transmission in one global \\
			& round for UE $n$ associated with SBS $s$\\
			
			$t_{n}^{\textrm{com}}$ &  Achievable throughput when the UE $n$ is assigned to\\
			& a set of SBSs\\
			
			$T_s^{\textrm{global}}$ & All-group overall delay of UEs associated\\ & with server $s$ \\
			
			$k \in \{1,2,\ldots\}$ & Number of hierarchical levels\\
			
			$\mathscr{P}_n$ & Personalized learning problem\\
			
			$\mathcal{P}^l_s, \mathcal{P}^l_n$ &  Preference list of SBSs on a set of UEs $n$, and \\
			& of each UE over SBSs, respectively\\
			$ U_n(\cdot), U_s(\cdot)$ & Utility function of UE $n$, and SBS $s$, respectively \\
	\hline
	\end{tabular}
\end{table} 

In the following, we first give a brief summary of the learning mechanism involved in the Edge-DemLearn in Subsection \ref{sub:iii_a}, and define the distributed learning scheme in a group in Subsection \ref{sub:iii_b} leveraging the HetNet infrastructure with MEC-enabled SBSs. Next, we derive communication model in Subsection \ref{sub:iii_c}, followed by computation model  in Subsection \ref{sub:iii_d}, and finally transmission time measurements in Subsection \ref{sub:iii_e} to formulate a general optimization framework with the objective to minimize overall latency and support basic FA, as conceived in \cite{federatedanalytics} with FL.

\subsection{Overview of the Learning Processes Involved}\label{sub:iii_a}
For a given learning task, each UE in the cellular network can run multiple iterations over their available data samples to train a local learning model. Following the multi-connectivity scenario, these UEs can transmit such local models to the SBSs available in the coverage area. The MBS facilitates these associations between UEs and SBSs following some strategy (discussed in Section~\ref{sec: Solution Approach}) to perform model aggregation, and further executes hierarchical clustering mechanism to form logical learning groups based on the similarities in learning characteristics of UEs such as the model parameters, i.e., the gradients and weights. The UEs in each logical learning group will try to improve their personal learning model (the process named as \textit{Specialization}) while considering the attributes of generalized global model to minimize generalization errors (the process named as \textit{Generalization}), i.e., the model performance on unseen or new data samples. Note that the local model parameters of each UE in a particular logical learning group are firstly aggregated at the SBSs, and then the partial aggregated group learning models are forwarded to the MBS for all-group aggregation to construct regional learning model. Thus, this results in partial in-group model aggregation corresponding to their groups at the SBS. Whereas, at the MBS, the full model aggregation, i.e., inner-group and then intra-group is performed to build a generalized regional model. Hence, we observe aggregation being carried out at different levels in an edge-assisted hierarchical learning in HetNet topology, as shown in Fig.~\ref{F:Edge_DemAI}, which are primarily facilitated by multi-connectivity and distributed MEC\footnote{The MEC also acts as an analytics server.} infrastructures. While the system architecture is flexible enough to leverage distributed computing nodes like MEC servers for distributed aggregation, the central idea is to improve the overall learning performance (i.e., better specialization and generalization) to an accuracy level.    

\subsection{Distributed Learning Scheme in a Group}\label{sub:iii_b}
We start with a general scenario of distributed learning where we consider each participating UE $n \in \mathcal{N}_{g \in \mathcal{G}}$ stores its local dataset $\mathcal{D}^g_n $ of size $D_n^g$. Then, we define the total training data size in each group $g\in \mathcal{G}$ as $D^g=\sum_{n=1}^{N}D_{n}^g$. In a typical supervised learning setting, $\mathcal{D}_n^g$ defines the collection of data samples given as a set of  input-output pairs $\{x_i, y_i\}_{i=1}^{D_n^g} $ in a group $g \in \mathcal{G}$. The feature values $x_i\in \mathcal{X}$ and the corresponding labels $y_i\in \mathcal{Y}$, where $\mathcal{X} \subseteq \mathbb{R}^d$ and $\mathcal{Y} \subseteq \mathbb{R}$, respectively. We define a predictor $F(w):\mathcal{X}\rightarrow\mathcal{Y}$ for some $w\in \mathbb{R}^d$. Note that the data samples can be generated through device's usage, such as interactions with applications. This results in statistical heterogeneity involving the non-i.i.d characteristics and unbalanced dataset. For the ease of representation, hereafter, we will drop the superscript $^g$ and consider the learning problem for each group as follows. 

With the sample data $\{x_i,y_i\}$, a typical learning problem for an input sample vector $x_i$ (e.g., the pixels of an image) is to find the  \textit{model parameter vector} $w_i\in \mathbb{R}^d$ that characterizes the output $y_i$ with the loss function $f_i(w)$. 

For each UE $n$, a general empirical loss with respect to $w$ on the local data set $\mathcal{D}^g_n$ is
\begin{equation}
	J_{n}(w|\mathcal{D}^g_n) := \frac{1}{D^g_n} \sum \nolimits_{i = 1}^{D^g_n}f_i(w_i) + \eta R(w_i),
	\label{eq:localloss}
\end{equation}	
where $\eta$ is a regularization coefficient and $R(w_i)$ is a regularization term which has several variants, such as $\sum\nolimits_i|\boldsymbol{w}-\boldsymbol{w}_i|$ and $\sum\nolimits_i||\boldsymbol{w}-\boldsymbol{w}_i||^2$ for L1 and L2 regularization, respectively, for any arbitrary iteration \cite{nguyen2020self, li2020federated}.
Then, a general group learning problem can be defined as the finite-sum objective of the form
\begin{equation}
\underset{w \in \mathbb{R}^d}{\text{min}}J_g(w) := \sum\nolimits_{n = 1}^{N^g} J_n (w|\mathcal{D}_n^g),
\label{eq:learning_problem}
\end{equation}
where $J_g(w)$ is the learning loss of group $g$, and $N_g$ is the number of group's members.

\textbf{\textit{Hierarchical Distributed Learning Scheme}:}
The learning problem in \eqref{eq:learning_problem} is solved following an iterative approach as follows:
\begin{itemize}
	\item In each $t^{\textrm{th}}$ update of global iteration, each participating UE $n$ implements stochastic gradient descent (SGD) over its on-device training data $\mathcal{D}_n$ to obtain local group learning models. These models are transmitted to the associated SBSs following a certain rule, i.e., the allocation of wireless communication resources.
	\item At the SBSs, the group parameters are evaluated, aggregated, and the partial aggregated group learning models are forwarded to the MBS for all-group aggregation to construct regional learning model. In particular, all of the local learning model parameters of group members distributed across the SBSs are aggregated at the MBS. Once the \textit{in-group} averaging is performed, at MBS, \textit{intra-group} aggregation (a.ka. hierarchical averaging) is done and the regional learning model is broadcast back to all the corresponding UEs in the network for the next global iteration.  
	
\end{itemize}  
This process continues until a certain level of global accuracy level is obtained. Furthermore, we also observe the MEC does not access the local data $\mathcal{D}_n^g, \forall n$ for training a global model. Instead, it aggregates the local learning model parameters to obtain the global model parameters.

Next, in the following subsections, we will discuss the communication and computation cost models involved to realize learning models within groups by leveraging the edge-assisted democratized learning approach.

\subsection{Communication Model} \label{sub:iii_c}
Each UE $n \in \mathcal{N}$ performs reference signal received power (RSRP) measurements, based on which the serving MBS will manages connections between potential SBS and UE. Then, the achievable throughput from the assigned SBS $s$ is defined as 
\begin{equation}
    r_{n, s} = \xi_{n, s}\hat{M}_{RB}\log_2\bigg( 1 + \frac{P_s^tG_{n,s}}{\sum \nolimits_{s'\in S}P_{s'}^tG_{n,s'} + \sigma^2} \bigg),
    \label{eq: single_connection}
\end{equation}
where $\xi_{n, s}$ is the number of RBs assigned to the UE $n$ from SBS $s$, $P_s^t$ is the transmit power, $G_{n,s}$ is the channel gain, and $\sigma^2$ is the channel gain, respectively. Consider the case where the UE $n$ is assigned to a set of SBS as $\tilde{S}_n \subseteq \mathcal{S}$. Then, the overall throughput under such scenario can be defined as 
 \begin{equation}
     r_{n, \tilde{S}_n} = \sum \nolimits_{s \in \tilde{S}_n}\xi_{n, s}\hat{M}_{RB}\log_2\bigg( 1 + \frac{P_s^tG_{n,s}}{\sum \nolimits_{s'\in S}P_{s'}^tG_{n,s'} + \sigma^2} \bigg).
     \label{eq:multi_connection}
 \end{equation}
Assume that the number of subbands allocated to SBS $s$ is $\zeta_s$. Then, the RBs allocated to all UEs and SBSs should not exceed the total number of RBs in the system, which is given as the following constraint:
\begin{equation}
     \sum_{\forall n \in \mathcal{N}_{g\in \mathcal{G}}, \forall s \in \mathcal{S}} \xi_{n, s} \le \sum_{\forall s \in \mathcal{S}}\zeta_s = M_{RB}. 
\end{equation}
Considering \eqref{eq: single_connection} and \eqref{eq:multi_connection}, we can generalize the achievable throughput of UE $n$ in terms of uniform resource allocation to each SBS $s \in \mathcal{S}$ as
 \begin{equation}
     r_{n} = \sum \nolimits_{s \in \tilde{S}_n}\beta_n^s\zeta_s \hat{M}_{RB} \log_2(1+ \gamma_{n,s}),
     \label{eq:general_connection}
 \end{equation}
where $\gamma_{n,s} = \frac{P_s^tG_{n,s}}{\sum \nolimits_{s'\in S}P_{s'}^tG_{n,s'} + \sigma^2}$, and $\beta_n^s$ is the fraction of bandwidth allocation for device $n$ from SBS $s$. Here, a single SBS contribution is captured when $|\tilde{S}_n|=1$. Hence, taking multi-connectivity into account instead of traditional fixed connectivity approach, UEs can exploit a number of radio links to find a suitable one that improves the overall model aggregation time in a democratized learning paradigm. In fact, this is in line with the requirements of the next generation machine learning applications where reliability should be considered while accelerating the knowledge acquisition process via edge computing infrastructures. 

\subsection{Computation Model}\label{sub:iii_d}
As in a general FL setting, the complex dependency between the number of global iterations, the number of local iterations, and the mini-batch size during local training can largely impact the trained global model accuracy and the model convergence rate \cite{FL_advances}, \cite{federatedSurvey2020}. Similarly, based on the nature of learning tasks, several existing works aim to quantify the relationship between the number of local computations, the local accuracy, and the required global iterations to reach an accuracy level. In this regard, as an example, for a general convex machine learning task, the number of global iterations, i.e., the communication rounds between the UEs and the parameter aggregator (MEC sever), is lower-bound in relation with local accuracy $\theta$ and global accuracy $\epsilon$ as \cite{konevcny2016federated}:
 \begin{equation}
     I^{\textrm{global}}(\epsilon, \theta) =  \frac{\delta\log(1/\epsilon)}{1-\theta},
     \label{eq:global_iterations}
 \end{equation}
where $\delta$ is some constant that depend on the nature of the learning task. Furthermore, the number of iterations each UE $n$ takes to obtain small local accuracy $\theta$ is given as
 \begin{equation}
     I^{\textrm{local}}(\theta) =  \nu_n\log\bigg(\frac{1}{\theta}\bigg),
     \label{eq:local_iterations}
 \end{equation}
 where $\nu_n$ is a parameter choice of UE $n$ that depends on the data size and condition number of the local learning problem. Then, using these definitions, the total computation time per global iteration at each UE $n$ is given by 
  \begin{equation}
    t_n^{\textrm{comp}} = I^{\textrm{local}}(\theta) \times \frac{c_n D_n}{f_n},
     \label{eq:local_time}
 \end{equation}
 where $c_n$ accounts for the total number of CPU-cycles and $f_n \in [f_n^{\textrm{min}}, f_n^{\textrm{max}}]$ denotes the allocated CPU-frequency of UE $n$, respectively.

\subsection{Transmission Time Measurements}\label{sub:iii_e}
Given a fixed local model parameter data size at UE $n \in \mathcal{N}$ as $d_n$, the expected per-global-round communication latency during parameter transmission when UE $n$ is associated with SBS $s$ is defined as
  \begin{equation}
    t_{n,s}^{\textrm{com}} = \frac{d_n}{r_{n,s}}.
     \label{eq:global_time_MC}
 \end{equation}
Note that initially UE $n$ can be primarily associated to any of the SBS $s\in  \mathcal{S}$, provided RSRP measurements at the MBS, and can belong to any of the logical learning groups.  
Similarly, we can derive the achievable throughput via multi-connectivity as 
  \begin{equation}
    t_{n}^{\textrm{com}} = \frac{d_n}{r_{n}}.
     \label{eq:global_time_MC}
 \end{equation}
As a starting point to do latency analysis, we first explore the latency measurements between UE $n$ and a single SBS $s$ following traditional single-connectivity assumptions, whereas multi-connectivity is exploited when performing SBSs-UEs optimal assignments. To that end, considering asynchronous aggregation at the MBS, the all-group overall delay associated with local computation at UE $n \in \mathcal{N}^s$ and its communication with the aggregation server $s$ can be equivalently derived as 
  \begin{equation}
   T_s^{\textrm{global}} = I^{\textrm{global}}\max \{t_{n,s}^{\textrm{com}} + t_n^{\textrm{comp}} \}, \forall n \in \mathcal{N}^s.
   \label{eq:global_time}
 \end{equation}
Here, we omit a possible extra delay term inside $\max\{\cdot\}$ assumed to capture the delay during communication between the SBSs and MBs, and the hierarchical model aggregation. In fact, it is intuitive to eliminate this constant measurement in further analysis as the latency at MBS during higher layer aggregation (i.e., model averaging) is very less, given high-speed fiber connectivity between the SBSs and MBS and available better computing capacity. Besides, we limit the scope of evaluation of system-level heterogeneity at the UEs level, i.e., UEs have heterogeneous computing abilities, similar to several existing works in distributed machine learning over wireless \cite{pandey2020crowdsourcing,tran2019federated,le2020incentive}. In this regard, considering a general optimization framework to minimize overall FL latency at layer-1 of SBS $s$, we have the problem defined as
	\begin{mini!} [2]                
		{\boldsymbol{\tilde{\beta},\tilde{f}}}                               
		{T_s^{\textrm{global}}}{\label{opt:P1}}{\textbf{P1: }}   
		\addConstraint{\sum \nolimits_{n \in \mathcal{N}^s}\beta_n^s \le 1 ,}{ \label{cons:fraction_resource}}
		\addConstraint{0 < \beta_n^s \le 1,}{\; \; \forall n \in \mathcal{N}^s, \label{cons:resouce_limiation}}
		\addConstraint{f_n \in [f_n^{\textrm{min}}, f_n^{\textrm{max}}],}{\; \; \forall n \in \mathcal{N}^s \label{cons:local_frequency}},
		\addConstraint{\cup_{g \in \mathcal{G}}\mathcal{N}_g^s = \mathcal{N}^s,}{\; \; \mathcal{N}^s \subseteq \mathcal{N} \label{cons:total_UEs}},
		\addConstraint{\mathcal{N}_i^s \cap \mathcal{N}_j^s = \emptyset,}{\; \; \forall i,j \in \mathcal{G} \; \text{and} \; i \neq j\label{cons:UE_each_group}}.
	\end{mini!}
In \eqref{opt:P1}, the uplink communcation constraints and the local computation capacity are captured, respectively, with \eqref{cons:fraction_resource}--\eqref{cons:resouce_limiation} and \eqref{cons:local_frequency}. Constraint \eqref{cons:total_UEs} ensures that all associated UEs in each logical learning group will perform the model training, and \eqref{cons:UE_each_group} requires that one UE cannot belong to more than one logical learning group to perform local model updates and transmission.

In this work, we will not focus on a joint energy consumption and learning performance optimization problem in FL. Rather, we adopt a mechanism to handle fast knowledge acquisition following the democratized learning framework, as explained in the following subsection. Note that posing joint energy consumption and learning performance optimization problem will not alter gist of the discussion presented in this work. Moreover, besides a novel hierarchical structure to support FA, in this paper, we concentrate more on improving generalization via Dem-AI systems, and leave additional details of FA as future work. 

\section{A framework for Dem-AI systems at the network edge}\label{sec:framework}

In this section, we propose the deployment design of the Dem-AI systems with MEC servers, as illustrated in Fig.~\ref{F:Edge_DemAI} in Section \ref{sec:system_model}, for a single common learning task of all UEs. We first define learning process in Dem-AI systems in Subsection \ref{sub:iv_a}, and the synergy of resource allocation and learning in Subsection \ref{sub:iv_b}, followed by a general formulation for an edge-assisted democratized learning   in Subsection \ref{sub:iv_c}. For the ease of representation and analysis of the main proposal of this paper, i.e., edge-assisted democratized learning framework towards FA, we resort limiting the proposed general framework, as in Fig.~\ref{F:Edge_DemAI}, to a 2-level structure with a regional MBS and associated SBSs, which we have explained in the Section \ref{sec:system_model}. Hence, in a region, the MEC server manages the collaborative learning process and knowledge integration from groups and learning UEs. Whereas, at the MBS, we perform regional level aggregation related with overall in-group and intra-group models. Besides, given the UEs exchanging only the model parameters with the MEC servers and not their raw data, the approach is privacy-preserving compared to conventional machine learning at a central entity (or remote cloud) with UEs' local data transferred. And different from two steps of aggregation in client-edge-cloud hierarchical FL \cite{liu2019client}, we propose updates in MEC servers at the SBS and regional aggregation at the MBS; hence, we unleash a novel infrastructure for FA at different levels.

\subsection{Learning Process in Dem-AI Systems}\label{sub:iv_a}
In this subsection, we first present a high-level architecture design of a Dem-AI system. Then, we present the synergy of resource allocation and learning problem. For simplicity, we remove discussions on the underlying network topology which facilitates self-organizing hierarchical learning mechanism via multi-connectivity in this subsection (we refer to Subsections \ref{sub:iii_d} and \ref{sub:iii_e} for this). 

At the lowest levels of the hierarchical generalized model construction in Dem-AI systems, the regional learning consists two generalized levels. At each level $k \in \{1, 2, \ldots\}$ in Dem-AI, the generalized model $\boldsymbol{w}^{(k)}$ is constructed as follows: 	
   \begin{align}
	\boldsymbol{w}^{(k)}
	=\sum\nolimits_{g \in \mathcal{G}} \frac{N_{g}^{(k-1)}}{N_g^{(k)}}\boldsymbol{w}^{(k -1)}_g,  (\emph{Hierarchical Averaging}), \label{EQ:H_averaging}
	\end{align}
where $N_{g}^{(k-1)}$ is the number of UEs in logical learning group $g$, and $N_{g}^{(k)}$ is the total number of UEs in the current specialized group at level $k$. Here, at the personalized level $0$, $N_{g}^{(0)} = 1$. In regional learning, the actual learning process can only be performed at the UEs by using their available local personal data. Thus, at the lowest level $0$, the UE $n$ will solve its personalized learning problem\footnote{We obtain personalized learning models solving this. In fact, these models are tailored for individual UE after hierarchical knowledge transfer at different aggregation levels (i.e., at SBSs and the MBS). We refer our readers to \cite{nguyen2021distributed} for further details on how we build personal models, the model training process, and other intricate mechanisms of Dem-AI systems.} $\mathscr{P}_n$ by incorporating the generalized learning models from higher-level groups and the personalized learning objective. Therefore, $\mathscr{P}_n$ in hierarchical learning structure is formulated as \cite{nguyen2020self}:
    \begin{align}
    \label{EQ:LPL_level_0}
    \mathscr{P}_n
	:=\underset{\boldsymbol{w}\in \mathbb{R}^d}{\text{min.}}~ J^{(0)}_n (\boldsymbol{w}|\mathcal{D}^{(0)}_n) + \eta\sum\nolimits_{k=1}^{K}\frac{1}{N_g^{(k)}}\|\boldsymbol{w}-\boldsymbol{w}^{(k)}\|^2 ,  
    \end{align}
where $J^{(0)}_n$ is the learning loss function of the learning UE $n$ for a classification task given its personalized dataset $\mathcal{D}^{(0)}_n$ and the learning model ${w}^{(k)}$ of the higher-level groups. Intuitively, the higher levels of generalized knowledge are less important than the lower-level specialized knowledge to solve the personalized learning tasks of UEs. 

    
The learning UEs in the region will learn and periodically send the updated learning information such as learning models and gradients towards the MEC servers at SBSs over the access networks. At the SBSs, the aggregation operation for each group in the region follows the averaging in (\ref{EQ:H_averaging}) to build the group generalized knowledge (i.e., $k=1$). Since only a subset of members is associated with each SBS, MEC servers at SBSs will perform partial aggregation of group models. Thereafter, SBSs send the aggregated partial models to the MEC server at the MBS to perform full averaging for each group, and then aggregate them to build a regional generalized model (i.e., at the level $k=2$). The regional model is broadcast to all users (or optionally can be first sent to the central cloud for another level of hierarchical aggregation, and then broadcast back to all users) to perform another update, as in Fig.~\ref{F:Edge_DemAI}. In addition to the regional model, the higher level generalized models are also updated and incorporated in the personalized learning at UEs. However, the update frequency for the model in the region is more frequent than the update in the cloud since the regional knowledge is also more helpful for each user than the higher-level of generalized knowledge from the cloud. Therefore, we focus on the knowledge acquisition mechanism within region.

\subsection{Synergy of Resource Allocation and Learning }\label{sub:iv_b}

The resource allocation problem can be formulated independently as in Subsection \ref{sub:iii_e}; however, edge-assisted Dem-AI system would have synergy benefits using novel designs for the communication and learning problem. In particular, considering the trade-offs between the learning performance and resource utility objectives can efficiently improve the learning performance of the regional model under the limited regional resource constraints. Moreover, learning groups compete with each other for communication resources to accommodate large group members, ensure frequent model updates, and thereby improve the group learning performance. 
 However, the formulated problem is challenging due to the complex fusion of existing resource allocation schemes, the usability of heterogeneous access networks, the locations of learning group members among different SBSs, and the coupling learning performance of groups and regional model.
 	
  The regional learning objective based on their group learning objective components at level-$2$ can be defined as \cite{nguyen2020self}:
   	\begin{align}
 	\underset{\boldsymbol{w}^{(2)}, \boldsymbol{w}^{(1)}_1,\dots,\boldsymbol{w}^{(1)}_{|\mathcal{S}_K|}}{\min} &  \sum\nolimits_{i \in \mathcal{S}_K} \frac{N_{g,i}^{(1)}}{N_g^{(2)}}\Big( {J}_{i}^{(1)}(\boldsymbol{w}^{(1)}_i|\mathcal{D}^{(1)}_{i})+ \frac{\eta}{2} \|\boldsymbol{w}^{(1)}_{i} - \boldsymbol{w}^{(2)}\|^2\Big),
 	\label{EQ:GK1}
	\end{align}
 where $\mathcal{S}_K$ is the set of subgroups of the top level group, ${J}_i^{(1)}$ is the loss function of subgroup $i$ given its collective dataset $D_{i}$, and $N_g^{(2)} = \sum\nolimits_{g \in \mathcal{G}} N_{g,i}^{(1)}$ . The objective function is weighted by a fraction of the number of learning UEs $N_g^{(1)}$ of the subgroup $i$, and the total number of learning UEs $N_g^{(2)}$ in the system. Hence, the subgroups which have more number of learning UEs have higher impact to the generalized model at level-$2$.
 \subsection{Edge-assisted Democratized Learning: a General Formulation}\label{sub:iv_c}
  	Given the $G$ learning groups, the challenging problem is: \textit{How can we make a joint decision on the user association}, and \textit{how can we utilize the communication resource for exchanging the learning models to build the regional generalized knowledge more efficiently?} In doing so, minimizing the overall delay of the aggregation operation process becomes one of the primary objectives to speed up the regional learning process. In particular, the regional model at the MBS can be solely constructed when all groups finish their aggregation that depends on maximum delay of the multiple aggregation points in the set of SBSs $\mathcal{S}$ in a multi-connectivity scenario. Similarly, at the MBS, each group aggregation can only complete after receiving personalized models from all members who are associated with the corresponding SBSs. 
Hence, it pose a problem of joint resource allocation amongst UEs in the proposed 2-layer HetNet architecture. Furthermore, these joint decisions can be made by solving the problem given the predetermined local training time of UEs based on their learning capabilities and the number of gradient-based updates of personalized learning problem. 

\textbf{\textit{Joint Association and Resource Allocation Design in Heterogeneous Networks Towards FA}}:
Following the above discussions, we redefine the overall all-group global learning time in the 2-layer HetNet architecture as  
       \begin{equation}
          T^{\textrm{global}} = I^{\textrm{global}}\max \{t_{n}^{\textrm{com}} + t_n^{\textrm{comp}} \}.
             \label{eq:overall_global_time}
        \end{equation}
Thus, the optimization problem is formulated as
   	\begin{mini!}[2]                
		{\boldsymbol{\beta, f}}                               
		{T^{\textrm{global}}}{\label{opt:P3}}{\textbf{P2: }}   
		\addConstraint{\sum \nolimits_{n \in \mathcal{N}^s}\beta_n^s \le 1,}{\; \; \forall s \in \mathcal{S}, \label{cons2:fraction_resource}}
		\addConstraint{0 < \beta_n^s \le 1,}{\; \; \forall n \in \mathcal{N},\; \forall s \in \mathcal{S}, \label{cons2:resouce_limiation}}
		\addConstraint{f_n \in [f_n^{\textrm{min}}, f_n^{\textrm{max}}],}{\; \; \forall n \in \mathcal{N} \label{cons2:local_frequency}},
		\addConstraint{\cup_{g \in \mathcal{G}}\mathcal{N}_g^s = \mathcal{N}^s, \cup_{s \in \mathcal{S}}\mathcal{N}^s = \mathcal{N}}{ \label{cons2:total_UEs}},
		\addConstraint{\mathcal{N}_i^s \cap \mathcal{N}_j^s = \emptyset,}{\; \; \forall i,j \in \mathcal{G},\; \text{and} \; i \neq j, \forall s \in \mathcal{S} \label{cons2:UE_each_group}}
		\addConstraint{\mathcal{N}_i^s \cap \mathcal{N}_i^{s'} = \emptyset,}{\; \; \forall s, s' \in \mathcal{S}. \label{cons2:UE_each_sbs}}
	\end{mini!}
Following \textbf{P1}, the updated constraint \eqref{cons2:total_UEs} ensures all UEs associated with logical learning groups in the network will perform the model training. Constraints \eqref{cons2:UE_each_group} and \eqref{cons2:UE_each_sbs}, respectively, requires that one UE cannot belong to more than one logical learning group to perform local model updates and transmission to multiple SBSs at a time. Moreover, we observe optimization problem \textbf{P2} reveals a viable and more practical infrastructure other than basic single server-multiple clients architecture of FL towards FA. This unravels a new direction towards distributed database management by exploiting quality of model at different levels and regions. To that end, we also realize \textbf{P2} offers a flexible mechanism for fast knowledge acquisition that can be leveraged to derive statistical insights of distributed datasets, which is the objective of FA.

For a given number of UEs associated with a SBS, the resource allocation problem \textbf{P1}
is a convex optimization, which can be solved using the Karush-Kuhn-Tucker (KKT) conditions, similar to \cite{tran2019federated}. However, the problem \textbf{P2} is hard to solve due to the coupling of logical association constraints \eqref{cons2:total_UEs} and \eqref{cons2:UE_each_group} with allocation of computation and communication resources in the objective. Furthermore, the combinatorial nature of the problem, which is generally NP-hard, makes it challenging to establish a robust, self-organizing learning architecture envisioned for the next generation of applications. In this regard, considering the suitability of matching theory to overcome such challenges, we adopt a two-sided many-to-one matching algorithm \cite{roth1992two}, which produces feasible sub-optimal solutions with low-complexity. In particular, we leverage three fundamental benefits of matching theory \cite{gu2015matching, han2017matching}: (i) ability of decentralized operations that can capture complex network components, (ii) distributed resource management with guarantees of stability, and (iii) self-organized, low-complexity and sub-optimal solutions, to fulfill the objective of Edge-DemLearn. In the following section, we present the details of our solution approach.
    \begin{figure}[t!]
        	\centering        	
        	\includegraphics[width=0.7\linewidth]{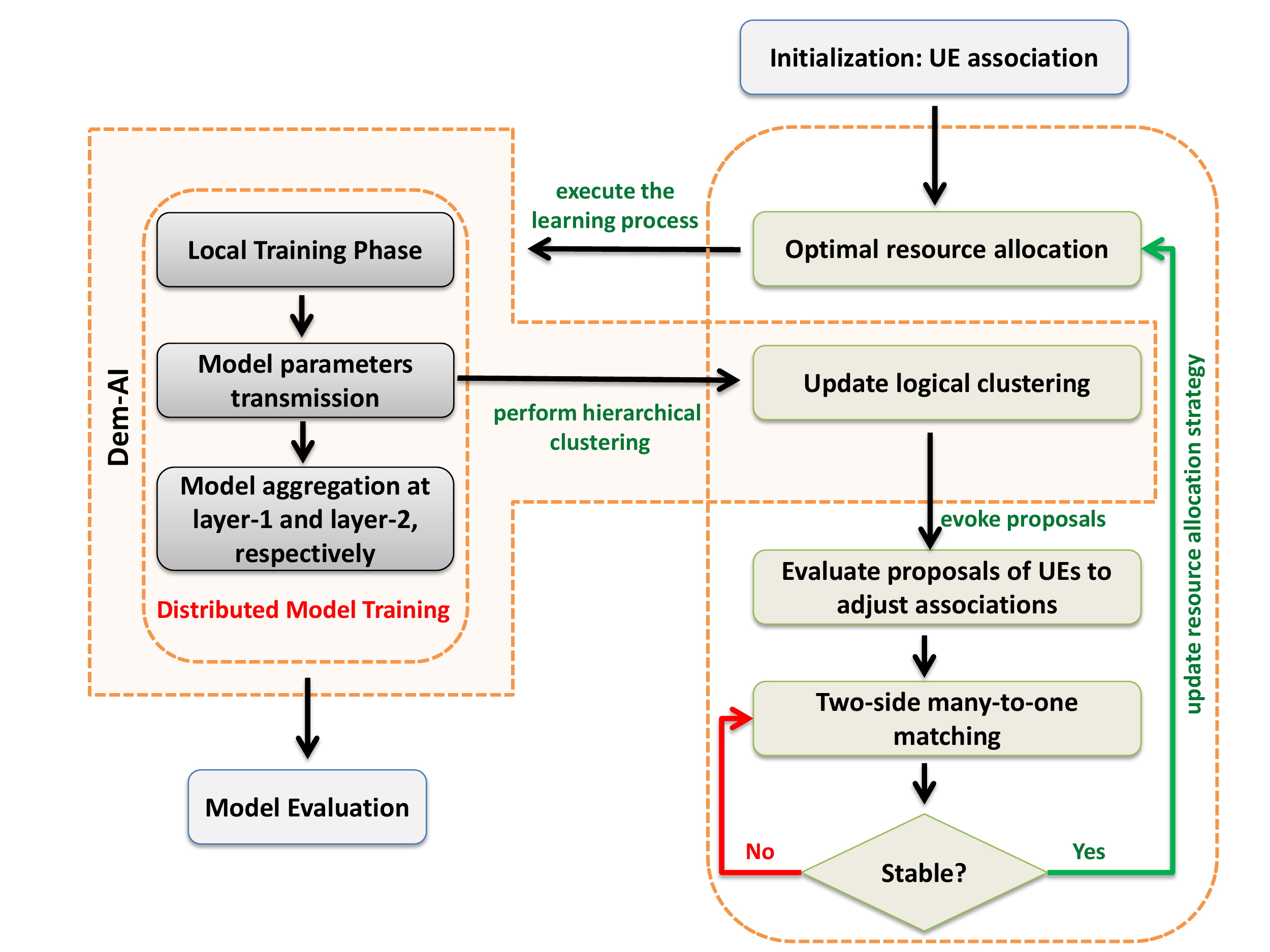}
        	\caption{An illustration processes in a single round execution of edge-assisted democratized learning framework, \textit{Edge-DemLearn} for a 2-layer HetNet architecture.}
        	\label{fig:illustration}
    \end{figure}  
    
\section{Proposed Solution Approach} \label{sec: Solution Approach}

We handle combinatorial properties of \textbf{P2} following an iterative two-sided many-to-one matching algorithm with externalities. In doing so, we first consider the learning problem at each SBS. To start with, we consider initial association of UEs with SBSs is based upon the RSRP levels of each UE. Then, we separate problem \textbf{P2} as \textbf{P1} and solve it for each SBS $s$. Given the problem \textbf{P1} as a convex optimization problem for a number of associated UEs $n\in\mathcal{N}^s$, we can obtain $\{\beta^*, f^*\}$ for the joint learning and resource optimization problem following the decomposition and solve technique. For this, we first find the optimal solution $f^*$ keeping the association fixed.

For the given association between SBSs and UEs, and known bandwidth allocation, the optimal solution is derived based on the following definition.
\begin{definition}\label{def1}
The optimal solution \textit{\textbf{f}}$^*$ for the minimization problem under fixed bandwidth allocation is $f_n^{\textrm{max}}, \forall n \in \mathcal{N}^s$.
\end{definition}
\begin{proof}
The objective of the minimization problem \textbf{P2} considers the knowledge aggregation time only, instead of focusing on the local energy constraints as well. Thus, we achieve this boundary solution to the constraint on \textit{\textbf{f}} for such kind of problems. Furthermore, as we have discussed earlier, the analysis of the proposed framework remains unaffected given the joint energy-learning formulation in the optimization problem\footnote{We consider this extension as our future work.} for the optimal solution of \textit{\textbf{f}}, as in \cite{tran2019federated}. 
\end{proof}

Next, given the solution $f^*$, we find out $\beta^*$ using the popular ECOS solver in the toolkit CVXPY \cite{diamond2016cvxpy}. In what follows, we execute the Dem-AI process and employ hierarchical clustering based on local model parameters, i.e., weights and gradients, to organize UEs in multiple logical learning groups. To that end, a two-sided many-to-one matching is applied to improve the knowledge acquisition process during model parameters aggregation at the SBSs by associating appropriate set of UEs in multiple logical learning groups under each SBS. In doing so, we exploit the interactions between UEs in different logical learning groups to handle the overall network-wide latency. This means, the UEs can leverage multi-connectivity and dynamically update their preference profiles to minimize their model aggregation latency with associated SBSs, whereas the SBS will also update its preference lists to improve the overall utility value which is defined in terms of parameter aggregation time.
The MBS mediates this process to obtain a stable matching solution for solving \textbf{P2}. The details of our proposed solution is illustrated in Fig.~\ref{fig:illustration}.

In the following, we present a near-optimal two-sided many-to-one matching solution in Subsection \ref{subsec:matching} to efficiently solve \textbf{P2}, followed by Edge-DemLearn algorithm and analysis in Subsection \ref{subsec: edge_demlearn}.

\subsection{A Two-sided Many-to-One Matching Solution}\label{subsec:matching}
We adopt a two-sided many-to-one matching game \cite{gale1962college, roth1992two} comprising the two disjoint set of players: the UEs $n\in\mathcal{N}$, and the SBSs plus local edge nodes, i.e., $s\in\mathcal{S}\cup  \{0\}$ such that $\mathcal{N}\cap\{\mathcal{S}\cup\{0\}\}=\emptyset$. Here, $\{0\}$ defines the set of local edge nodes which can be considered as a virtual SBS, with resource seen as infinity, and can serve each UE $n$ with a defined cellular resource to ensure stability. The game is played iteratively in each global round of training. We consider each SBS can communicate with a number of UEs based upon their substitutable preferences\footnote{With arbitary preferences a stable matching may not exists. Thus, we use substitutable preferences.} and designated ``quota" $q_s$, which its the overall resource capacity, however, each UE is associated with only one of the SBS. 
In the following, we formally define the many-to-one matching game for our proposed solution approach.
\begin{definition}
A many-to-one matching $\mu$ is a mapping from the set $\mathcal{N} \cup \mathcal{S}\cup\{0\}$ into the set of all subsets of $\mathcal{N} \cup \mathcal{S} \cup\{0\}$ such that for each $n \in \mathcal{N}$ and $s \in \mathcal{S}\cup\{0\}$ the following condition holds:
\begin{enumerate}
    \item $\mu(s) \subset \mathcal{N}$ and $\mu(n) \subset \mathcal{S}$.
    \item $|\mu(s)| \le q_s, \forall s \in \mathcal{S}\cup\{0\}$.
    \item $|\mu(n)| = 1, \forall n \in \mathcal{N}$.
    \item If $n \in \mu(s)$ for SBS $s$, then $s =\mu(n)$.
\end{enumerate}
where $q_s$ denotes the RB quota of SBS $s$.
\end{definition}
The constraints 2) and 3) ensures that the matching between SBS and UEs does not exceed SBSs resource capacity, and no two UEs are associated to a single SBS, respectively.

In this kind of matching game, each player (UEs and SBSs) has a preference profile over the other player (SBSs and UEs), i.e., in the formulated game, the SBSs $s, \forall s\in\mathcal{S}$ have preferences on a set of UEs $n$ and each UE has preferences over SBSs, initially defined as the RSRP levels\footnote{Note that the UE $n$ associated with SBS $s$ can belong to any of the logical learning group for performing model training. However, the concept of groups is only related with model aggregation during hierarchical knowledge transfer. Thus, we remove discussion on this notion in the hierarchical matching game.}, denoted respectively as $\mathcal{P}^l_s$ and $\mathcal{P}^l_n$. In our problem, the UEs can exploit multi-connectivity and switch its association between SBSs to minimize the model aggregation time for fast knowledge acquisition. This is particularly done to avoid ``stragglers" issues in the performance of the proposed learning system. To be specific, the preference of UEs to specific SBS may change as per the other UEs association to that SBS. Therefore, we draw the following conclusion.
\begin{remark}
The proposed matching game is a many-to-one matching with externalities.
\end{remark}
In the initial random matching, a UE $n$ ranks SBSs based on the observed RSRP, while the SBS $s$ ranks UEs $n, \forall n \in \mathcal{N}$ as per the incurred latency of model parameter transmission. Later, we adopt preference lists over the matching states following the utility function of UE $n$ as
\begin{equation}
    U_n(\mu(n,\beta^*)) = -T_s^{\textrm{global}}(\mu (n,\beta^*)), \forall n \in \mathcal{N}_g^s, \forall s \in \mathcal{S}, \label{eq:UE_utility}
\end{equation}
where $T_s^{\textrm{global}}(\mu (n, \beta^*))$ is the latency experienced by UE $n$ during parameter aggregation give the matching strategy $\mu$. 
Furthermore, the SBS formally defines its utility for UE $n$ as 
\begin{equation}
    U_s(\mu(n, \beta^*)) = -\sum\nolimits_{n \in \mathcal{N}}T^\textrm{global}(\mu(n,\beta^*)). \label{eq:SBS_utility}
\end{equation}
Then, the preference relation $\succ_s$ for any SBS $s$ defined over the set of UEs $\mathcal{N}_s$ can be defined as follows:
\begin{equation}
   (n, \beta^*) \succ_s (n', \beta^*) 	\Leftrightarrow U_s(n', \beta^*) >
   U_s(n, \beta^*), \forall n, n' \in  \mathcal{N}_s.
\end{equation}

In the formulated game, UEs are heterogeneous, and when they are matched to SBSs based on RSRP levels only, it will impact the overall model aggregation time at the regional level. Furthermore, the UEs may want to propose a change in their association with the SBSs to minimize the experienced latency at their end. Consequently, ``approvals" to such proposals may force the SBS to change its preference towards UEs as well. Thus, we define the network-wide association profile $\mathcal{A} = \{\mathcal{N}_s : \forall s \in \mathcal{S}\}$ as the set of UEs under multiple logical learning groups, performing local training, and transmitting model parameters to the SBSs. Following the preference list $\mathcal{P}_s^l$, which is updated when a certain number of global iterations is performed, the SBSs evaluate the association profile $\mathcal{A}$ to finalize the stable association strategy.

Following the concept of swap matching, we have the Definitions \ref{def3} and \ref{def4}, respectively, to define swap blocking pair and the two-sided stable matching.
\begin{definition}\label{def3}
A swap blocking pair under the matching $\mu$ is the pair of player $\{n, n'\} \in \mathcal{N}$ having concurrent different preferences over the existing matching solution, i.e.,  $n \succ_s n'$ for some $n' \in \mathcal{N}$ and $s \succ_n s'$ for some $s, s' \in \mathcal{S}$, respectively. Thus, we draw the following conclusion for a swap blocking pair under the matching $\mu$:
\begin{itemize}
    \item $\forall m \in \{n,n',s,s'\}, U_m(\mu_n^{n'}) \ge U_m(\mu),$ 
    \item $\exists m \in \{n,n',s,s'\},\; \textit{such that}\; U_m(\mu_n^{n'}) > U_m(\mu).$
\end{itemize}
\end{definition}
\begin{definition}\label{def4}
The matching $\mu$ is said to be stable if there exist no swap blocking pair, i.e., for a given association profile $\mathcal{A}$, we have $U_s(\mathcal{A}') > U_s(\mathcal{A}), \forall s \in \mathcal{S}$, given $\mathcal{A}'$ is the association profile other than $\mathcal{A}$.
\end{definition}
Following Definition \ref{def3} and Definition \ref{def4}, we can obtain optimal resource allocation variables $\{\beta^*, f^*\}$ for stable matching $\mu$ with the association profile $\mathcal{A}^*$ such that $ U_s(\mathcal{A}_-) > U_s(\mathcal{A}^*), \forall s \in \mathcal{S}, \forall \mathcal{A}_- \neq \mathcal{A}$.
\subsection{Edge-DemLearn: Algorithm and Analysis} \label{subsec: edge_demlearn}
The proposed solution approach operates in two-phases: \textbf{Phase I}, in which two-sided many-to-one matching with externalities is executed for optimal association between SBSs and UEs, and \textbf{Phase II}, in which Dem-AI learning mechanism is performed. In Fig.~\ref{fig:illustration}, we show an overview of the processes involved in a single round execution of edge-assisted democratized learning framework, \textit{Edge-DemLearn} for the proposed network architecture. To that end, we propose Algorithm~\ref{algo:Edge-DemLearn} based on two-sided many-to-one matching for implementing the processes illustrated in Fig.~\ref{fig:illustration}.
\begin{algorithm}[t!]
		\caption{Edge-DemLearn: Edge-assisted Democratized Learning } 
		\label{algo:Edge-DemLearn} 
		\begin{algorithmic}[1]
            \State \textbf{Input:} $k, \tau, \Lambda= 0, \mathcal{S}, \mathcal{N}, \mathcal{A}^0, \{ \mathcal{P}_n^l, \mathcal{P}_s^l, q_s, \mathcal{N}^s \}, \forall s \in \mathcal{S}.$
            
            \State \textbf{Initialization:} Random UE association with SBSs, $\mathcal{A}^0.$
            \For{$t \in 1, 2, ...,I$}
            \State \textbf{Phase I:}  Two-sided many-to-one matching for association and allocation between SBSs and UEs.  
            \While{$\Lambda == 0$}
        		\For{SBSs $s=1,\dots, S$}
        		    \State \multiline{
        		    Each SBS evaluates preference proposals $\mathcal{P}_n^l$ received via MBS for association with UEs.}
        		    \State \multiline{
        		    The UEs are ranked based initially subject to $\mathcal{P}_s^l$, given initial association profile $\mathcal{A}^{0}$ and $U_s(\mathcal{A}^0)$.}
        		    \State \multiline{
        		    Update $\mathcal{A}^{(t)}$ based on $U_s(\mathcal{A}^{(t)})$ such that $U_s(\mathcal{A}^{(0)}) > U_s(\mathcal{A}^{(t)})$.}
        		    \State Evoke ``proposals" from UEs:
            		   \While {Any updates in $\mathcal{P}_n^l$}
                		   \State \multiline{
                		   SBS evaluates $\mathcal{P}_s^l$ to adjust $\mathcal{A}^{t}$ for optimal $U_s(\mathcal{A}^{(t)}).$}
              		   \EndWhile
              		\State Set $\Lambda$ to  $1$. 
                \EndFor
            \State \multiline{
            Obtain stable matching $\mu$ for $\mathcal{A}^*$ resulting in $\{\beta^*, f^*\}$.}
            \EndWhile
            \State \textbf{Phase II:} Dem-AI learning mechanism. 
            \If {$(t \mod \tau = 0)$} 
            \For{UEs $n \in \mathcal{N}$}

            \State \multiline{
            Receive and update local model from the higher-level generalized models as \eqref{EQ:GK1}.}
            \State \multiline{
            Implement gradient based optimization to iteratively solve personal learning problem using higher-level generalized models $\boldsymbol{w}^{(k)}_n$ as \eqref{EQ:LPL_level_0}.}
    		 \State \multiline{
    		 Then, each UE $n$ sends updated learning model to the server. }

    		 \EndFor
    		 \EndIf
        	\State \multiline{
        	At each generalized level $k$ perform an update for the learning model as \eqref{EQ:H_averaging}.} 
            \State \multiline{
            Server reconstructs the hierarchical structure by the clustering algorithm.} 
            \EndFor
		\end{algorithmic}
	\end{algorithm}	 

As aforementioned, based on the RSRP levels, MBS will handle a random association between SBSs and UEs at $t=0$, $\mathcal{A}^0$. Following to which, the UEs will perform the first round of distributed model training. Once this process is completed, the UEs will transmit local model parameters to their respective SBSs. Then after, the aggregation is performed in each levels of the networks, i.e., at the SBSs and the MBS, along with hierarchical clustering based upon the local model parameters at the MEC server. Such hierarchical clustering enables the formation of logical learning groups of UEs. To that end, Phase I of Algorithm~\ref{algo:Edge-DemLearn} handles the association between SBSs and UEs (line 5--17). In particular, UEs (an example, say $n' \in \mathcal{N}$) will evaluate the utility function \eqref{eq:UE_utility}, and to further improve it, (may) send ``proposals" to their associated SBSs for updating the current association strategy. The corresponding SBS will announce the proposals made by their associated UEs to the SBSs in their preference list $\mathcal{P}_{n'}^l, \forall n' \in \mathcal{N}$, respectively. Note that the communication between SBSs for possible association adjustment of UEs only requires to acknowledge the current preference instances of UEs, i.e., just send these proposals to the SBSs to which the UEs can connect. Each SBS after receiving such proposals first enumerate them to assess the possibility of adjustment by comparing the difference in overall latency for knowledge acquisition (using \eqref{eq:SBS_utility}) when $n'$ is matched to any other SBS except the currently matched one in place of $n$. In doing so, the feasibility and stability of swap blocking pairs are evaluated as per the Definition 3. The process operates until finding all swap blocking pairs for the received proposals, i.e., there exist no more swap blocking pairs (line 14) and the matching is stable (line 16).   
	
Once the stable matching is reached, the UEs undergo executing Dem-AI learning mechanism (line 19--line 27). First, UEs will perform rounds of local iterations over their data samples leveraging higher-level generalized model (line 21--line 22). In particular, the UEs will implement stochastic gradient descent method to solve a personalized local objective function and produce local model parameters. The UEs will transmit back these parameters to their corresponding SBSs for partial aggregation of group models, and then to the MBS for building regional generalized model, following Dem-AI learning process (as described in Section \ref{sub:iv_b}). The MBS further instantiates the hierarchical clustering for logical learning group formation and returns back the process to Phase I for the next round of execution of Edge-DemLearn. Accordingly, the process is iteratively implemented until a level of accuracy is obtained.

\section{Performance Evaluation}\label{sec:performance_evaluation}
\subsection{Simulation Settings}
We consider a typical supervised learning setting to compare the efficacy of our proposed solution approach, Edge-DemLearn, against the well-known federated averaging algorithm, shortly FedAvg \cite{mcmahan2017communication}, over wireless networks. In fact, FedAvg is a common comparison baseline that works well with heterogeneous UEs communicating with a single aggregation server in the wireless networks \cite{dinh2020federated,zaw2021energy}, i.e., in a typical multiple UEs-single server architecture. However, different from existing approaches, in this work, we develop a hierarchical learning structure with distributed aggregation and control methodologies at the SBSs and the MBS in a 2-layer heterogeneous network topology. Then, given the stable matching solution, we first evaluate the performance of trained model for three distinct learning metrics to support FA: (i) Regional, (ii) Client Specialization, and (iii) Client Generalization, on MNIST \cite{lecun1998gradient}, Fashion-MNIST  \cite{xiao2017fashion_mnist}, and benchmark dataset Federated Extended MNIST (FEMNIST) \cite{caldas2018leaf}. Regional metric is the evaluation of model performance at the MBS, i.e., a regional level in our system model. Client Specialization and Client Generalization, respectively, are the evaluations of the trained model in the UE's local test data only and the unseen data which is prepared collecting test data from all UEs in the region. We shuffle and divide the dataset to replicate its unbalanced and non-i.i.d. properties at each UE; data samples are drawn particularly from two labels out of ten for MNIST and Fashion-MNIST. Furthermore, we reserve $20\%$ of data samples for the evaluation the model accuracy with the learning rate as 0.0001 and $\eta = 0.001$ and $k = 2$, i.e., two-level hierarchical averaging. We use Python $3.7$ and Tensorflow $1.13.1$ as our simulation framework, and performed experiments on the environment with the following hardware specifications: CPU Intel Core $i7-7700K 4.20(GHz)\times 8$; RAM $32$(GBs); GPU GTX 1080Ti $11$(GBs).

 We set a single MBS with the maximum system bandwidth as $3$ MHz, carrier frequency of $2$ GHz\cite{simsek2019multiconnectivity}. Unless stated otherwise, we set $|\mathcal{S}| = 3 \times 3$, i.e., 3 SBS per sector. The maximum transmit power is set as 23 dBm, and shadow fading standard is 3 dB. The UEs are assume to follow Homogeneous Poisson Point Process distribution (HPPP) between $2$ m and $100$ m in the network radius of $100$ m\footnote{To simplify, we consider a small network topology.}. Furthermore, we define $f_n^{\max}$ is uniformly distributed in the range $1.0 \sim 2.0$ GHz. In order to evaluate our proposal system model, we choose a set of users in between $10 \sim 100$ and the local model size range from $1000.0$ to $8000.0$ (KB) with mean value around $2712.0$ (KB), and median is around $2250.0$ (KB). The details of simulation parameters is in Table~\ref{tab:table_simulation}.
 
 \begin{table}[t!]
 	\centering
 	\caption{Simulation parameters.}
 	\label{tab:table_simulation}
 	{
 		\begin{tabular}{ |c|c| }
 			\hline
 			Parameter & Value\\
 			\hline
 			Number of UEs $|\mathcal{N}|$ & $10 \sim 100$ \\
 			\hline
 			Number of SBS $|\mathcal{S}|$ & $2 \sim 10$ \\
 			\hline
 			Path loss & $128.1 + 37.6 \log(d), d[km]$\\
 			\hline
 			Maximum transmit power & $23$ dBm\\
 			\hline
 			Shadow fading standard deviation& $3$ dB\\
 			\hline
 			System bandwidth & $3$  MHz\\
 			\hline
 			Bandwidth of each RB ($\hat{M}_{RB}$) & 	$180$ kHz\\	
 			\hline
 			Number of sub-carriers per RB & $12$\\
 			\hline
 			Model sizes & $ x=1000.0 \sim 8000.0$ \\
 			& $\bar{x} = 2712.0, \tilde{x} = 2250.0$ (KB)\\
 			\hline
 			Thermal noise for $1$ Hz at $20^{\cdot}$C & $-174$dBm\\
 			\hline
 	\end{tabular}}
 \end{table}
\begin{figure}[t!]
    \centering
    \begin{subfigure}{\linewidth}
        \centering
        \includegraphics[width=0.45\linewidth]{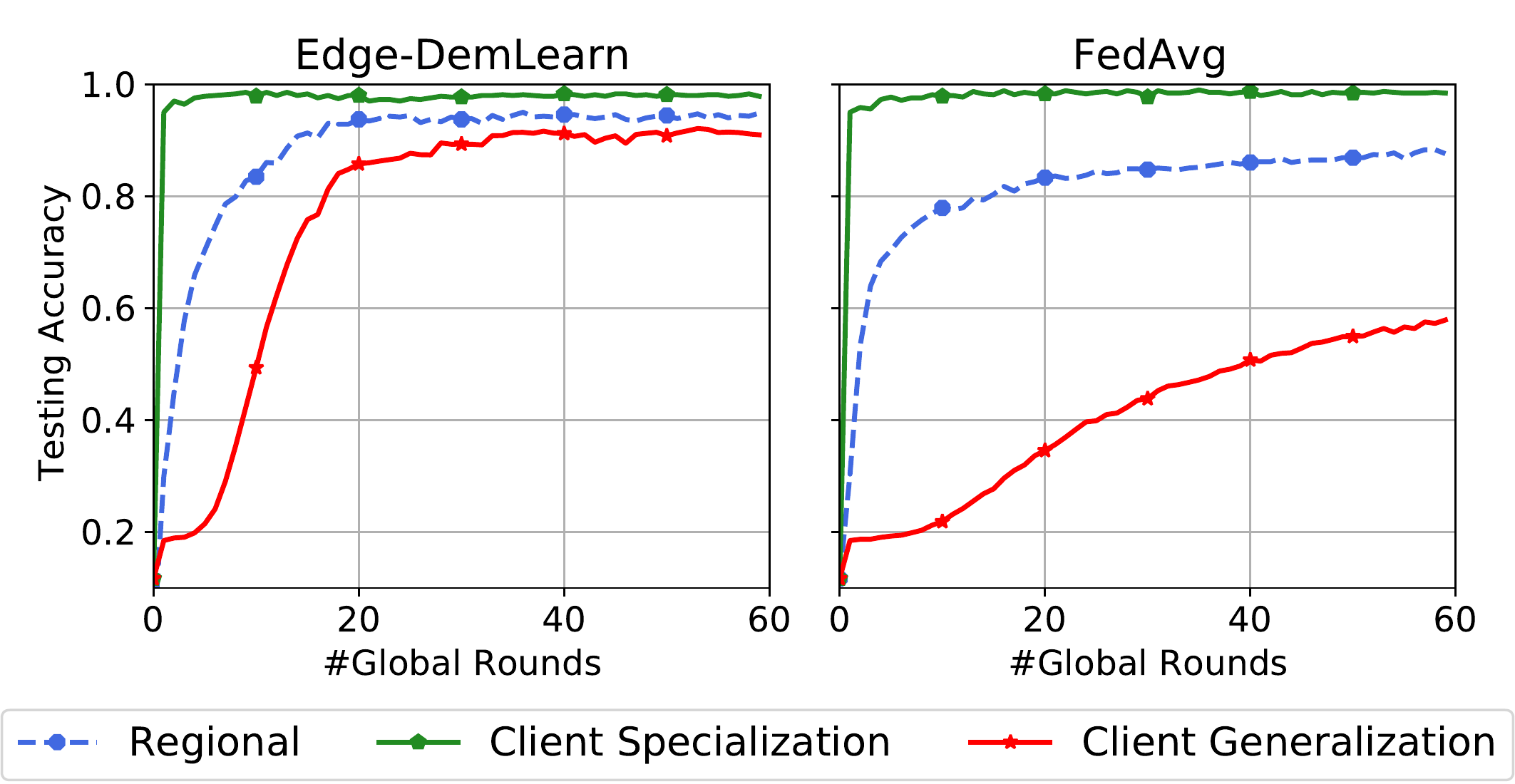}
    	\caption{Experiment with MNIST dataset.}
    	\label{F:DemLearn_a}
	\end{subfigure}
    \begin{subfigure}{\linewidth}
        \centering
        \includegraphics[width=0.45\linewidth]{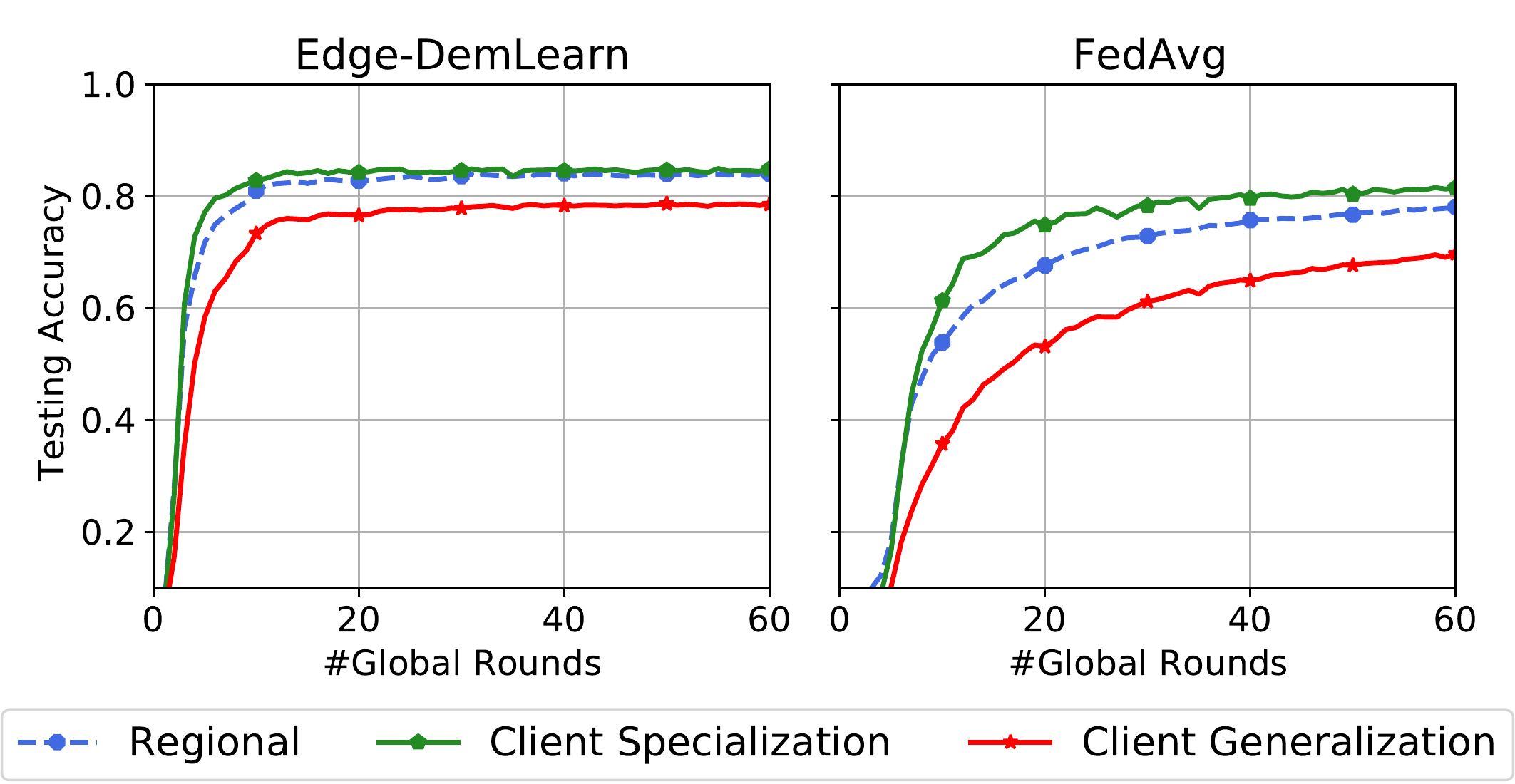}
    	\caption{Experiment with Federated Extended MNIST dataset.}
    	\label{F:DemLearn_c}
	\end{subfigure}
    \begin{subfigure}{\linewidth}
        \centering
        \includegraphics[width=0.45\linewidth]{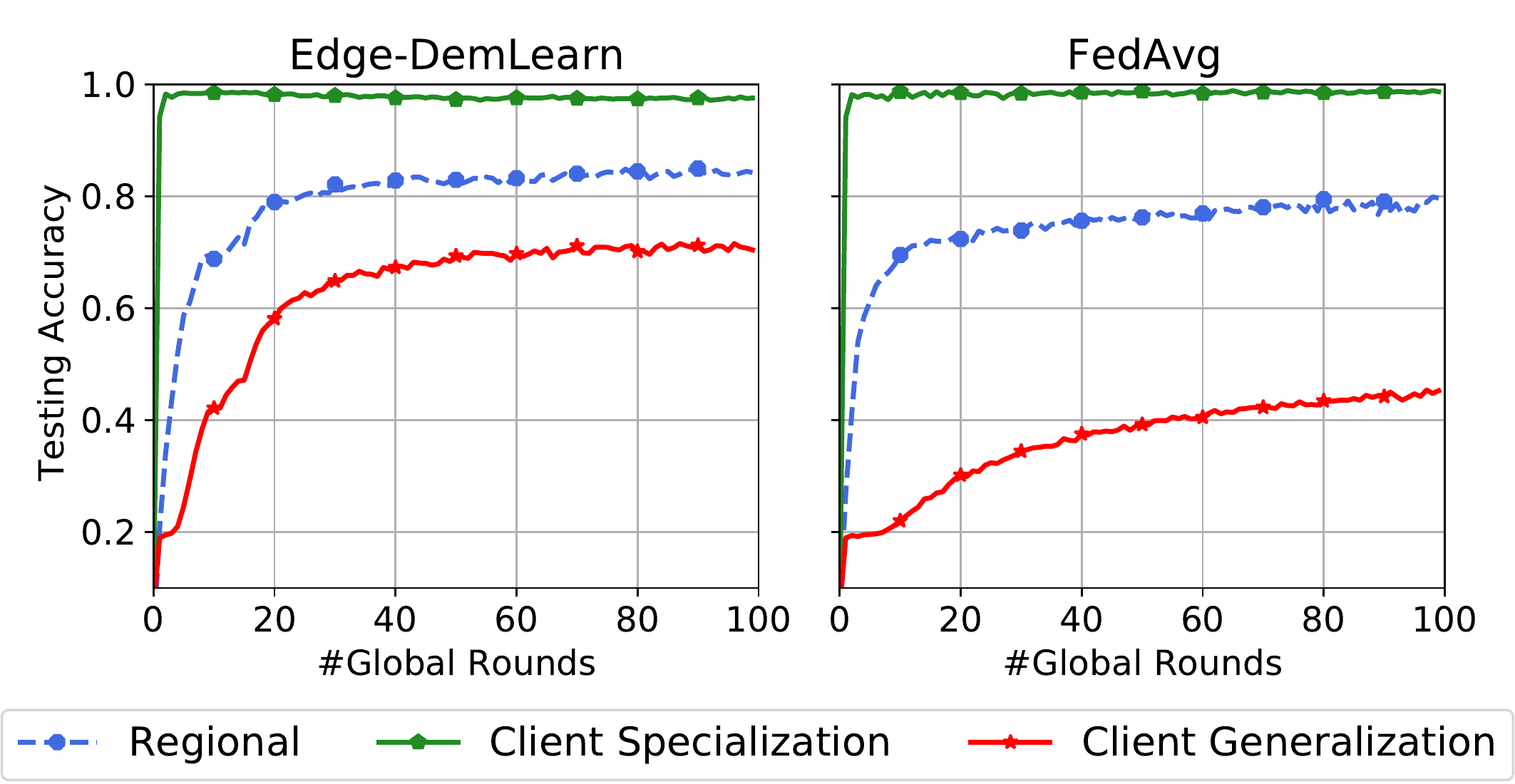}
    	\caption{Experiment with Fashion-MNIST dataset.}
    	\label{F:DemLearn_b}
	\end{subfigure}
	\caption{Learning performance comparison between Edge-DemLearn and FedAvg for $|\mathcal{N}|=$ 50 users.}
	\label{F:DemLearn}
\end{figure}

\subsection{Analysis of Model Performance and Comparisons}
In Fig.~\ref{F:DemLearn}, we show the regional learning performance for $|\mathcal{N}| = 50$ using MNIST, FEMNIST, and Fashion-MNIST dataset. Note that the model training time in Edge-DemLearn is evaluated with the number of global rounds required to attain a level of accuracy. This means we require more number of global rounds to obtain higher accuracy, and therefore, larger model training time. In particular, we obtain a running time of 0.00015s per set on average with our system's hardware configuration. In Fig.~\ref{F:DemLearn_a}, we observe Edge-DemLearn shows competitive performance against vanilla FedAvg in Client Specialization, whereas significantly outperforms FedAvg in terms of Client Generalization, which is critical metric in performing FA. Correspondingly, at 30 global rounds, we observe FedAvg obtains around $42\%$ test accuracy against $91\%$ for Edge-DemLearn in MNIST dataset. This produces fast knowledge acquisition for an improved learning model under less aggregation steps (or global rounds). As a result, we obtain a significant gain in communication cost due to parameter exchanges between UEs and the SBSs via wireless channels. Similarly, in Fig.~\ref{F:DemLearn_c} and Fig.~\ref{F:DemLearn_b}, we observe the proposed Edge-DemLearn outperforms FedAvg while ensuring significant improvement in Client Generalization.  

\begin{figure}[t!]
	\centering
	\includegraphics[width=0.35\linewidth]{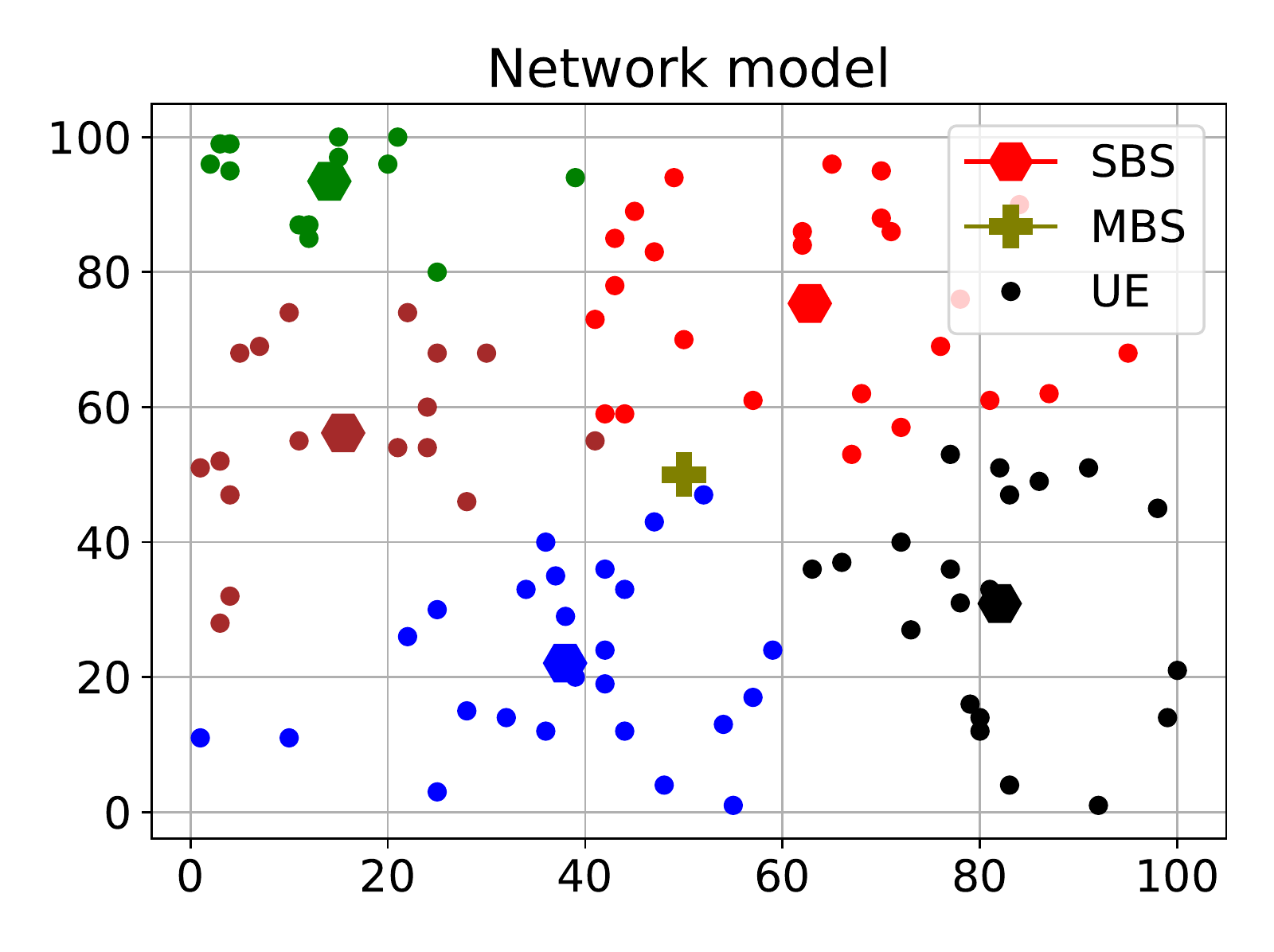}
	\caption{A snapshot of the network model.}
	\label{F:Network-model}
\end{figure}

\begin{figure}[t!]
	\centering
	\includegraphics[width=0.35\linewidth]{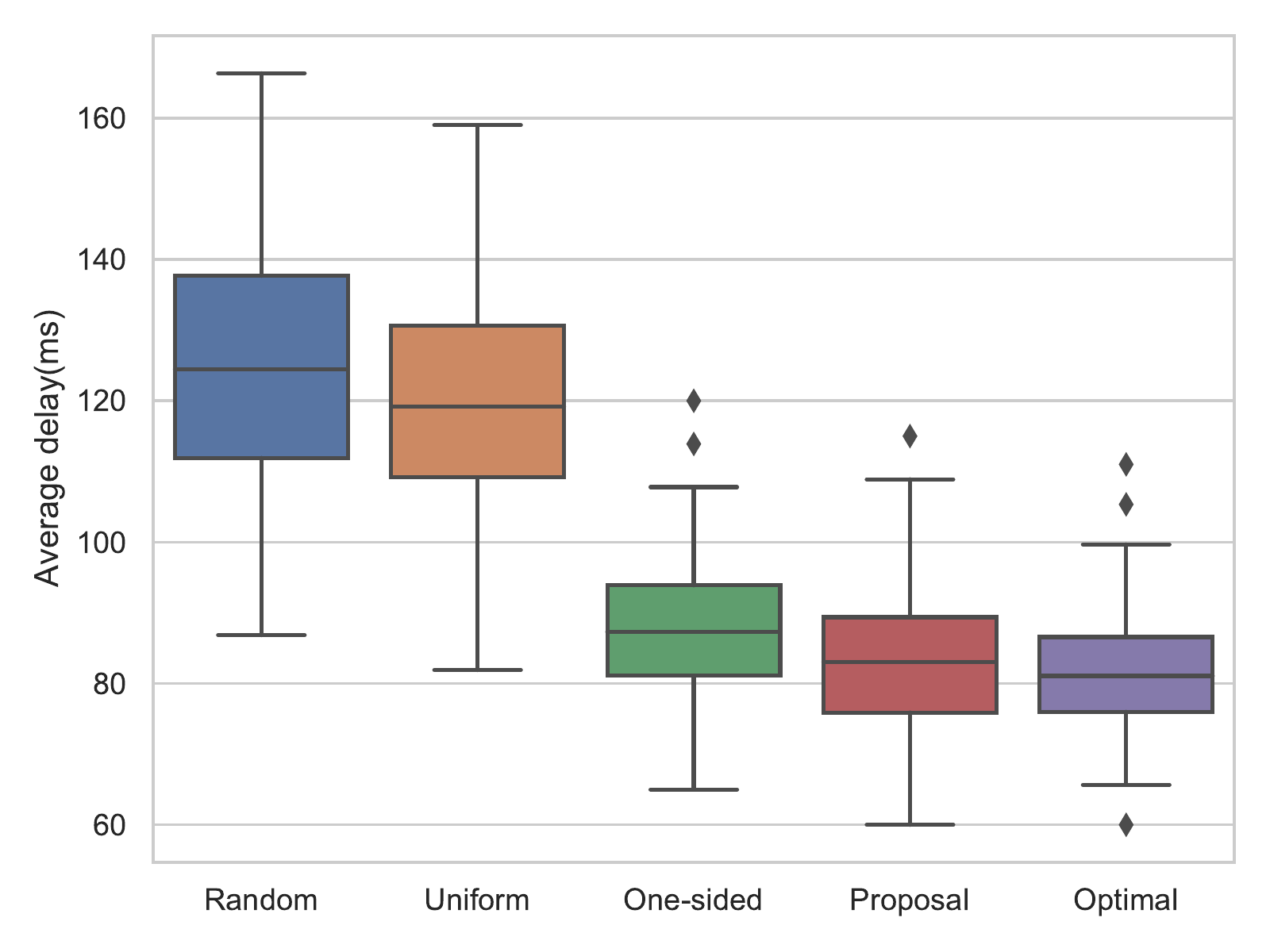}
	\caption{Total delay analysis.}
	\label{F:Matching}
\end{figure}

In Fig.~\ref{F:Network-model}, we present a snapshot of our network model. We set a network with the number of UEs at $N=100$ and the number of SBSs at $S=5$ using the aforementioned system variables and the simulations parameters in Table~\ref{tab:table_simulation}. Each ``circle" in Fig.~\ref{F:Network-model} denotes a UE, and UEs in color are associated with respective colored SBSs. The Fig.~\ref{F:Network-model} shows a double-sided many-to-one matching solution, similar in \cite{kazmi2017mode}, when realizing Edge-DemLearn algorithm. We observe the UEs are associated with respective SBSs in such a way that enhances the overall system utility while minimizing the learning time objective in the HetNet architecture.


In Fig.~\ref{F:Matching}, we evaluate the performance of double-sided matching solution defined in Algorithm \ref{algo:Edge-DemLearn} (Proposal) for 100 global rounds with four intuitive baselines: (i) One-sided \cite{gu2018joint}, which is a traditional one-sided stable matching that considers the preference of UEs as fixed in our settings, (ii) Random, which basically is a low-complexity approach that randomly associates a subset of UEs to the SBSs as per the distance between them, (iii) Optimal solution \cite{kazmi2017mode, papadimitriou1998combinatorial}, and (iv) Uniform, which is a naive approach that allocates equal fraction of resources to UEs associated with the respective SBSs (i.e., we divide the available bandwidth equally likely for each UE), in terms of total delay. We set a network with the number of UEs at $N=100$ and the number of SBSs at $S=5$. We then compare the results in terms of total delay (in \textit{ms}) required to obtain a certain accuracy level for the given global rounds. Moreover, the simulation results presented are derived from executing algorithms $50$ times and then averaged, to validate the obtained results' consistency. From Fig.~\ref{F:Matching}, we observe that our proposed scheme shows significant gain average $47.2\%$ as compared with the Random approach in terms of latency, $45.1\%$ with Uniform, $2.46\%$ with One-sided,  while avg. $1.8\%$ gap with the Optimal, respectively. The results are intuitive as Edge-DemLearn finds the best configuration and optimizes the network resources when associating the subset of UEs with SBSs. Furthermore, the proposed algorithm gives near-optimal solution, which is of low-complexity than the Optimal. Moreover, we observe high variance in the obtained solution using the Random and Uniform schemes. This is because they suffer from both random placement of UEs in the network following a HPPP and an inefficient allocation of network resources at each initialization. Also, note that One-sided scheme shows competitive performance in terms of overall delay; however, it cannot satisfy the objective of self-organizing hierarchical learning structure in support of FA because it restricts the preferences of UEs as fixed.

\begin{figure*}[t!]
	\centering
	\begin{subfigure}{0.31\textwidth}
		\centering
		\includegraphics[width=1\textwidth]{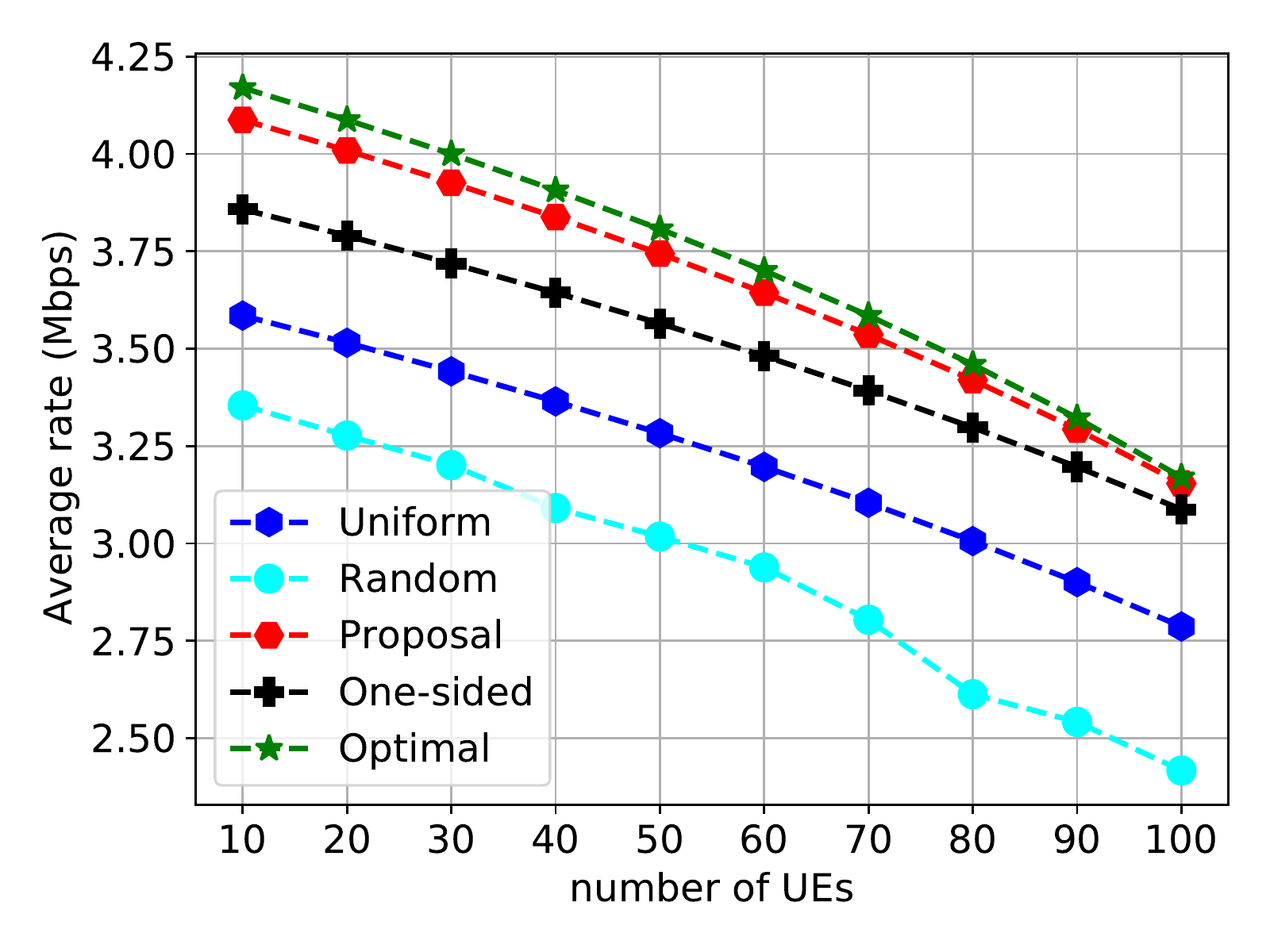}%
		\caption{}
		\label{Fig:data-rate}
	\end{subfigure}
	\hfill
	\begin{subfigure}{0.31\textwidth}
		\centering
		\includegraphics[width=1\textwidth]{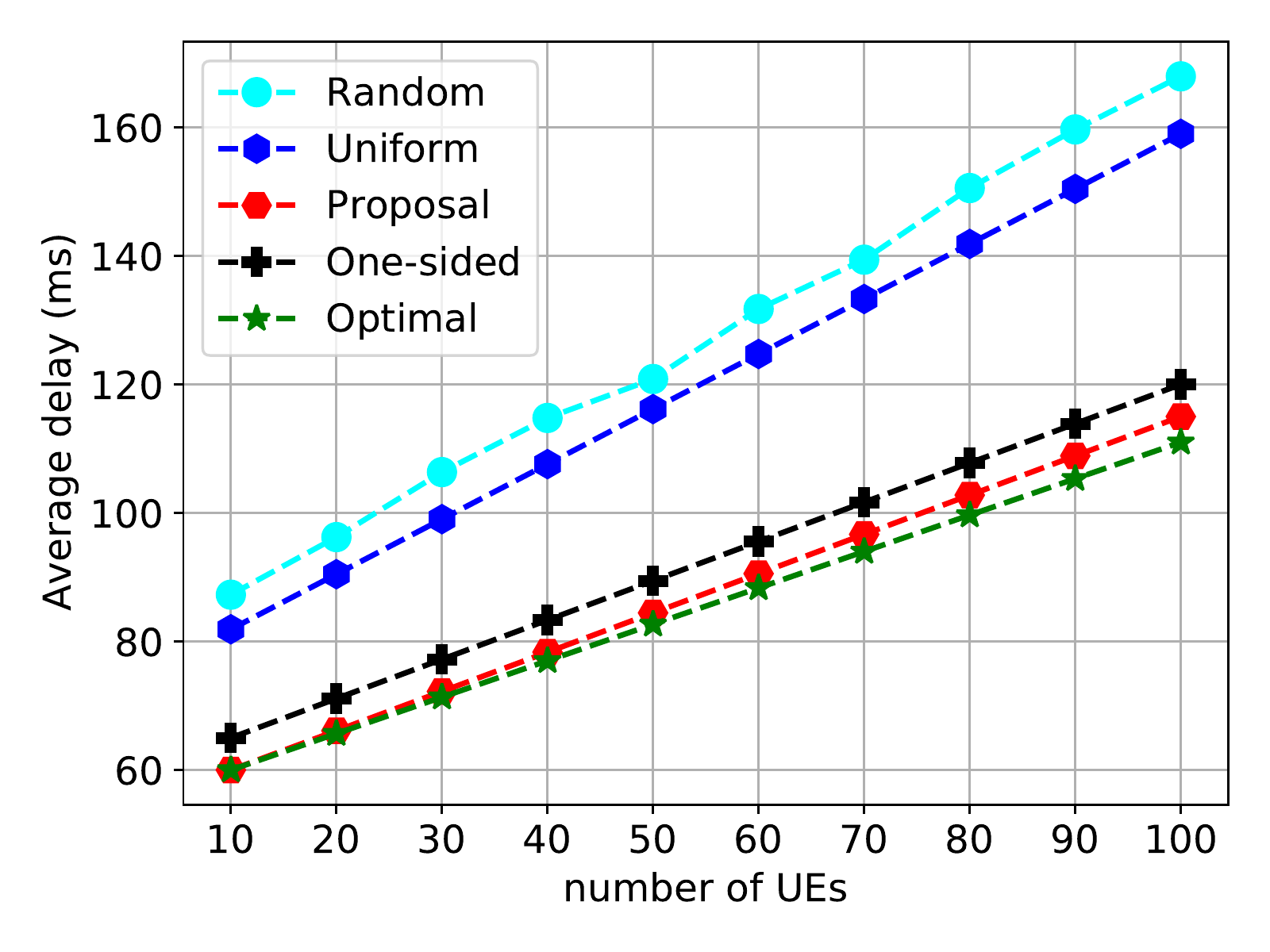}%
		\caption{}
		\label{Fig:avg-delay}
	\end{subfigure}
	\hfill
	\begin{subfigure}{0.31\textwidth}
		\centering
		\includegraphics[width=1\textwidth]{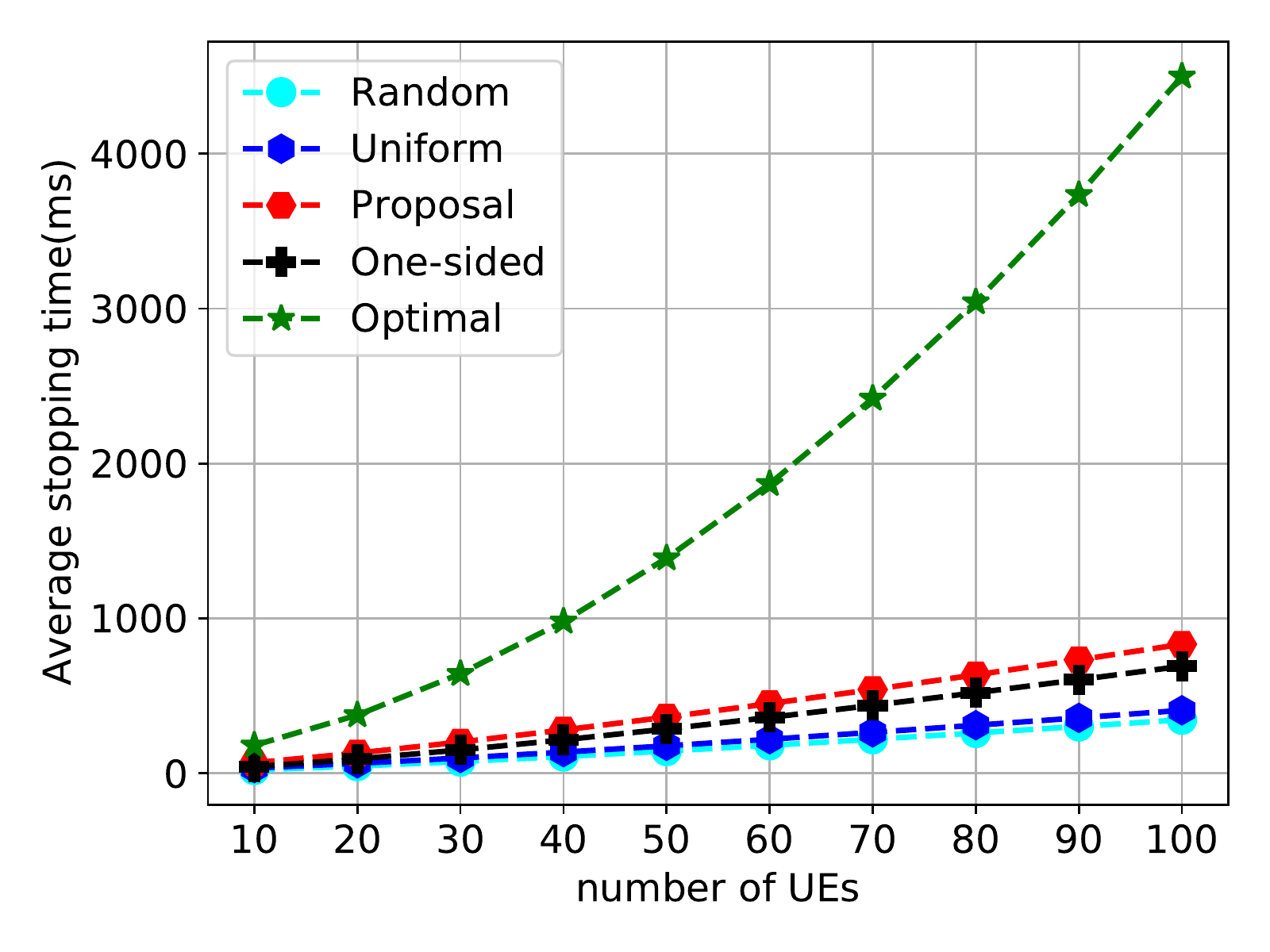}%
		\caption{}
		\label{Fig:avg-stop}
	\end{subfigure}
	\caption{Performance comparison between the Proposal, Random, Uniform, One-sided, and Optimal schemes with the number of UEs for a fixed number of SBSs $S=5$ in terms of: (a) average rate, (b) average delay, and (c) average stopping time.}
	\label{Fig:System_performance_matching}
\end{figure*}

\begin{figure}[t!]
	\centering
	\includegraphics[width=0.5\linewidth]{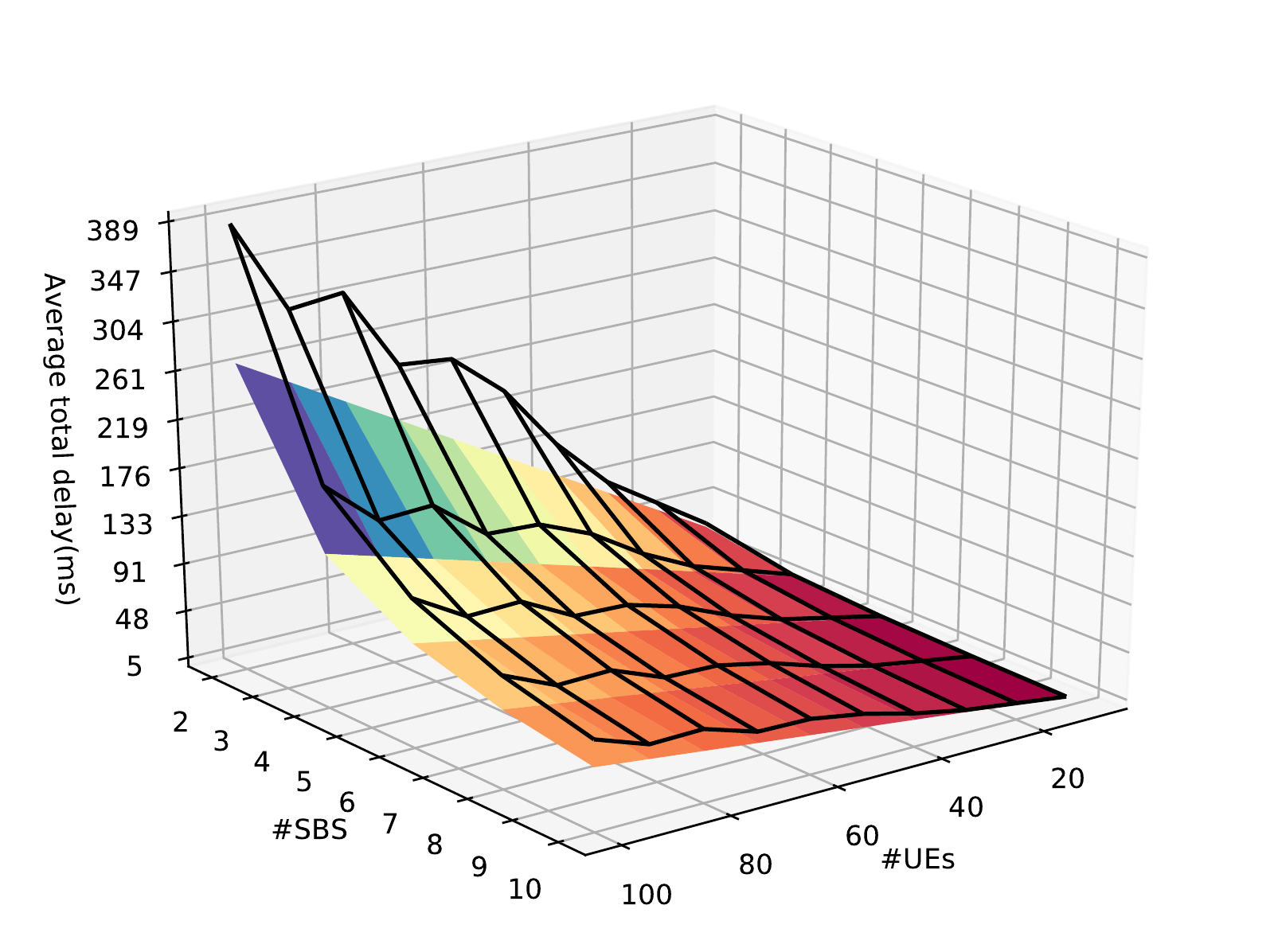}
	\caption{Qualitative analysis of average total delay vs. the number of SBSs vs. the number of UEs.}
	\label{F:Total_delay}
\end{figure}
Similarly, in Fig.~\ref{Fig:System_performance_matching}, we compare the proposed matching scheme's performance with baseline schemes while varying number of UEs for a fixed number of SBSs. Fig.~\ref{Fig:data-rate} shows the impact of the number of UEs on the achievable average rate. As expected, for a fixed number of SBSs, increasing the number of UEs introduces congestion in the wireless channel, and further causes interference; hence, we observe the impact of this phenomenon in the average rate. However, the proposed solution provides near-optimal solution compared with Uniform scheme which ignores the optimization problem and the Random that randomly associates UEs with the SBSs. Fig.~\ref{Fig:avg-delay} demonstrates the impact of the increasing UEs (i.e., the network density) on the average delay (in \textit{ms}). As illustrated, the average delay to converge increases linearly with the increase in the number of UEs. It is reasonable as the number of potential blocking pairs, i.e., $\binom{N}{2}+ N\times S$, increases linearly with the number of UEs $N$ \cite{roth1992two}. Here, $\binom{N}{2}$ defines the number of possible blocking pairs to swap with two SBSs and $N \times S$ is the number of  possible blocking pair to swap with one SBS and one available association. Moreover, we observe the proposed solution approach shows near-optimal solution and performs much better than Uniform scheme, which is quite intuitive. Finally, Fig.~\ref{Fig:avg-stop} illustrates the dependency of the number of UEs and the average stopping time (in \textit{ms}) which reflects the complexity of the proposed matching algorithm. We obtain a half-linear complexity of $\mathcal{O}(NS\log(NS))$ as the algorithm exploits interaction between UEs to optimize the system-level performance, i.e., matching with externalities. The Uniform scheme is a low-complexity solution but results in poor performance, as observed in Fig.~\ref{Fig:data-rate} and Fig.~\ref{Fig:avg-stop}, whereas the Optimal solution has exponential complexity  with the number of UEs and SBSs \cite{papadimitriou1998combinatorial}.

In Fig.~\ref{F:Total_delay}, we study the impact of the number of SBSs and the numbers of UEs (i.e., the network density) on average total delay (i.e., the system's utility). The heatmap is the Edge-DemLearn implementation for different settings of UEs and SBSs, wherein we demonstrate its comparative analysis against the Random baseline, as illustrated in Fig.~\ref{F:Total_delay}. We observe a significant gain in the total average delay (avg $18.7\%$) at high network density, i.e., $S=2, N=100$. In particular, we obtain $26.2\%$ better performance as compared with the baselines. The reason is that in a congested network scenario, the random association strategy restricts appropriate selection of UEs to improve overall knowledge acquisition time. Thus, resulting in more enormous delays. Furthermore, limited RBs hinder UEs' participation in an optimal way; hence, lowering the utility of the associated SBSs. Whereas the proposed algorithm allows SBSs and UEs to match in their best interest, thus improving the average total delay.

\begin{figure*}[t!]
	\centering
	\includegraphics[width=0.9\linewidth]{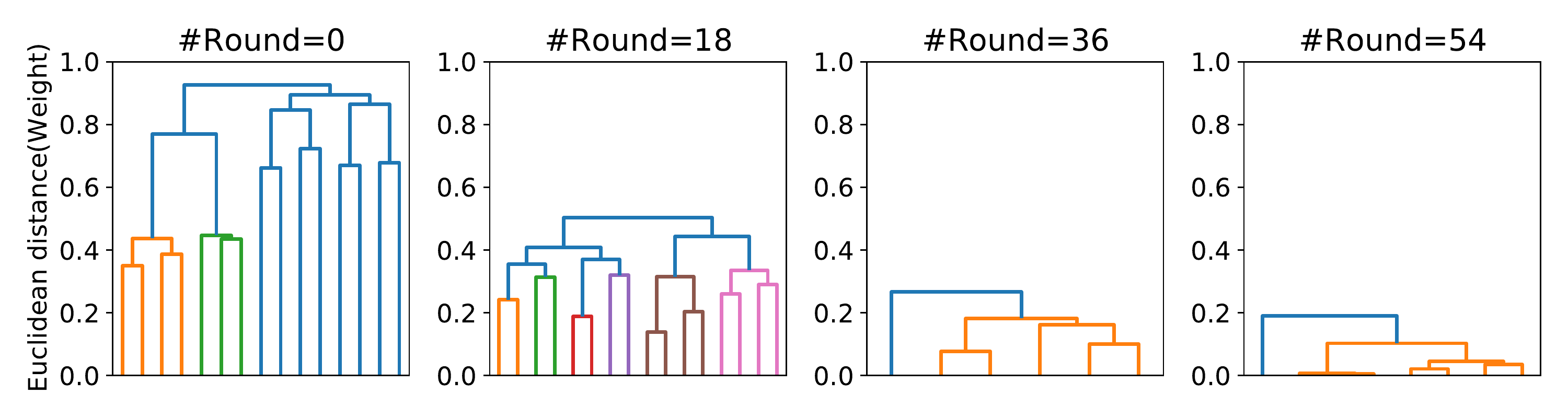}
	\caption{Snapshots of the logical clustering mechanism at different rounds. Left-Right: evolution of clusters throughout the model training process in Dem-AI.}
	\label{F:cluster}
\end{figure*}

Finally, in Fig.~\ref{F:cluster}, we provide a snapshot of the underlying logical hierarchical clustering mechanism at different rounds. Here, x-axis represents the groups to which UEs belong, and the y-axis represents the euclidean distance distance between model parameters (i.e., weights). The logical clustering is performed based on the euclidean distance between model parameters at the MBS. As illustrated in the figure, the hierarchical topology shrinks over the training time, indicating the convergence of the proposed Dem-AI learning mechanism. The impact of such change in the topology is observed in the model performance, where we obtain improvement in the Client Generalization (Fig.~\ref{F:DemLearn_a}, Fig.~\ref{F:DemLearn_b}). 

\subsection{Analysis of Data Privacy}
The hierarchical self-organizing structure of Edge-DemLearn requires UEs to share only the local learning information (i.e., the models and gradients) to their associated SBSs instead of the available raw datasets. To an extent, this facilities the data privacy of UEs; however, an adversary may reproduce the local training data samples, or its properties by exploiting the shared model information \cite{FL_advances} at different aggregation levels. Besides, apart from such adversarial inferences, leakages from aggregates also challenge the data privacy in Edge-DemLearn. In this regard, instead of averaging all local model parameters into a single model at the particular SBS, the Edge-DemLearn framework first performs partial in-group model aggregation (i.e., models are averaged based on their learning groups). Thus, it challenges differential attacks at the SBSs, and furthermore, can execute privacy-preserving functions of Differential Privacy (DP) \cite{wei2020federated,FL_advances} to mitigate privacy loss. Moreover, in the MBS, these partial aggregated in-group models undergo hierarchical averaging (i.e., in-group averaging considering the information received from the SBSs, followed by intra-group averaging) to build the regional model. In such a case, using a central DP method at the MBS, the Edge-DemLearn can protect UEs from adversarial attacks strongly. Following this, Edge-DemLearn can ensure the privacy of participating UEs' local data in the hierarchical model aggregation topology.




\section{Conclusion}\label{S:Conclusion}
In this work, we have proposed a novel edge-assisted democratized learning mechanism, namely \textit{Edge-DemLearn}, which unleashes a more practical hierarchical and distributed learning structure to support FA, while improving generalization capability (i.e., 95\%) of the trained model. In particular, the propose architecture leverages distributed MEC infrastructures at different regional levels to derive statistical insights on distributed datasets. To that end, we have developed a robust distributed control and aggregation methodology in regions by leveraging distributed MEC platforms in a HetNet environment. In doing so, we have adopted a multi-connectivity scenario to realize the hierarchical structure for FA, and applied a two-sided many-to-one matching algorithm with externalities to optimize the overall system performance during user associations and resource allocation. To that end, we have shown the proposed hierarchical structure enables fast knowledge acquisition (around 26.2\% improvement) to meet needs of both FA and model quality for future large-scale distributed learning systems. As compared with the prominent FL model training algorithm, such as FedAvg, using simulations on real datasets, we have shown the proposed approach can offer significant performance while minimizing the average total delays during knowledge acquisition.

\ifCLASSOPTIONcaptionsoff
\newpage
\fi
\bibliographystyle{IEEEtran}
\bibliography{Dem-AI}
\end{document}